\documentclass{wielgosz-info-article}

\usepackage{easy-todo}

\usepackage{siunitx}
\DeclareSIUnit{\deg}{deg}
\DeclareSIUnit{\pixel}{px}
\DeclareSIUnit{\fps}{fps}

\usepackage{tabularx}
\usepackage{nicefrac}
\usepackage{amsmath}

\author{Maciej Wielgosz}
\title{CARLA-BSP: a simulated dataset with pedestrians}
\papertype{Technical report}

\begin{document}
\begin{titlepage}
    \begin{authblk}{Maciej Wielgosz}{0000-0002-4401-2957}
        Centre de Visió per Computador (CVC), Barcelona, Spain
    \end{authblk}
    
    \begin{authblk}{Antonio M. López}{0000-0002-6979-5783}
        Centre de Visió per Computador (CVC), Barcelona, Spain
    \end{authblk}

    \begin{authblk}{Muhammad Naveed Riaz}{}
        Centre de Visió per Computador (CVC), Barcelona, Spain
    \end{authblk}

    \maketitle

    \begin{abstract}
        We present a sample dataset featuring pedestrians generated using the ARCANE framework, a new framework for generating datasets in CARLA (0.9.13). We provide use cases for pedestrian detection, autoencoding, pose estimation, and pose lifting. We also showcase baseline results. For more information, visit https://project-arcane.eu/.
    \end{abstract}
    
\end{titlepage}

\section{Introduction}
\label{sec:intro}

Realistic simulation frameworks can offer many advantages compared to real-world data. One of the most apparent is the instant availability of the ground truth, as opposed to the time-consuming manual labeling of the collected data. It is also relatively easy to generate a vast amount of samples, while the real-world data is often limited and, by necessity, unbalanced.

Autonomous vehicle (AV) testing and training is one example of an application where simulated data can provide an advantage. This boon is especially true when learn-from-data methods, such as deep learning (DL), are employed. Usually, the most dangerous situations occur rarely, and collecting enough examples to ensure the system handles them correctly can be a challenge -- because how often will you be able to record a kid playing with the dog beside the road?

Generation of such corner, adversarial cases is where simulation methods can help the most. However, even the most ,,typical'' outputs from the simulation can be of use -- imagine having an unlimited amount of data you can use to preliminary train your models or validate your flow.

CARLA (Car Learning to Act) is an open simulator for urban driving \cite{dosovitskiy17carla}. Among its other features, it offers a Python API that allows two-way interaction with the simulated world. One of the applications it can be used for is the creation of increasingly complex scenarios that can be recorded and utilized in various models' training and validation. Moreover, various modalities are available: RGB camera, DVS camera, LIDAR, and others. The ground truth annotations can be obtained, among others, using semantic and instance segmentation cameras or semantic LIDAR.

Of particular interest to us were scenarios involving pedestrians. Therefore, we decided to create a basic dataset containing simple scenes to validate the usefulness of this approach in practice and invite early community feedback. The dataset contains RGB camera data in the form of MP4 videos and semantic segmentation data encoded as APNG (Animated PNG) along with 2D and 3D pose keypoints. The videos involve a single pedestrian, with no other traffic data, only background objects (buildings, vegetation, parked cars). Four main pedestrian models are used: adult female and male, and children of both genders. Additionally, each of the basic models has look variants.

In the next sections, we describe the dataset in more detail and showcase its potential usage in various applications, briefly discussing state-of-the-art for those and similar tasks. The code is separated into two main repositories, the first one used to generate the dataset and the second one containing flows and models for common tasks. There are also helper repositories for orchestrating the whole setup and abstracting a common base. They are publicly available via \href{https://github.com/wielgosz-info/carla-pedestrians?utm_source=link&utm_medium=latex&utm_campaign=CARLA-BSP}{https://github.com/wielgosz-info/carla-pedestrians}. The dataset itself can be downloaded from \href{https://project-arcane.eu/datasets/binary-single-pedestrian?utm_source=link&utm_medium=latex&utm_campaign=CARLA-BSP}{https://project-arcane.eu/datasets/binary-single-pedestrian}.

\section{High-level concepts}
\label{sec:concepts}

Familiarity with notions presented in this section can help better understand the rest of this work. Some useful CARLA-related terminology and concepts, overall project idea, and code architecture are briefly described, along with some specific terms we use throughout the text.

\subsection{CARLA}
\label{ssec:concepts:carla}

\begin{figure}
    \centering
    \includegraphics[width=0.75\textwidth]{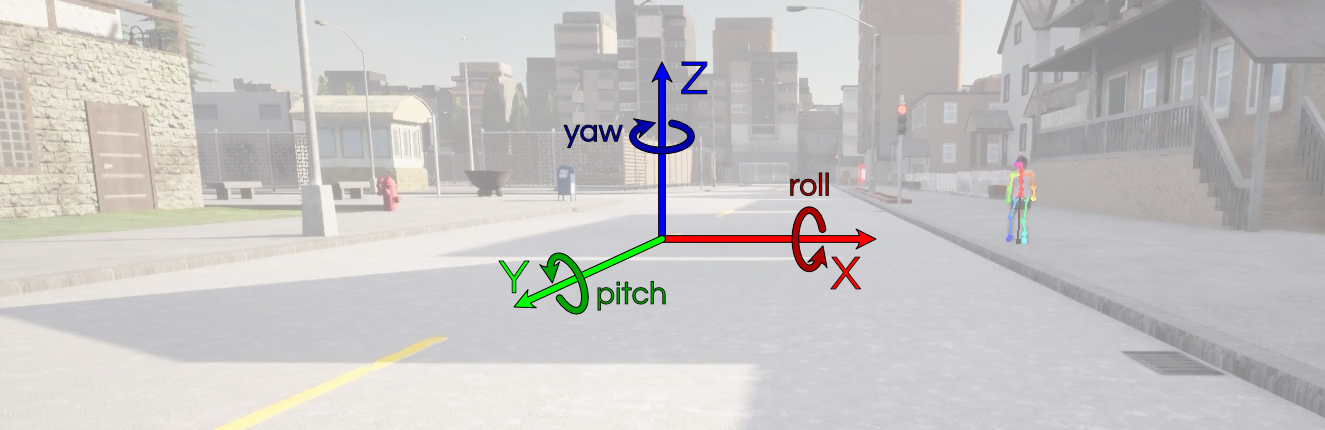}
    \caption{Coordinate system used by CARLA. \label{fig:concepts:carla:coordinate_system}}
\end{figure}

One crucial concept that must be introduced is the coordinate system. CARLA uses a Z-up left-handed system, as presented in Fig.~\ref{fig:concepts:carla:coordinate_system}. One of the basic structures that are used when working with the API is Transform, which consists of a Location (an X, Y, Z point in space) expressed in meters [\si{\meter}] and a Rotation (pitch, yaw, roll; note that the order does not correspond to axes order!) that is expressed in degrees. All of the following coordinates are expressed using this system unless otherwise stated.

\begin{figure}
    \centering
    \includegraphics[height=0.42\textheight]{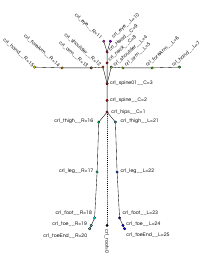}%
    \hspace{12pt}%
    \includegraphics[height=0.42\textheight]{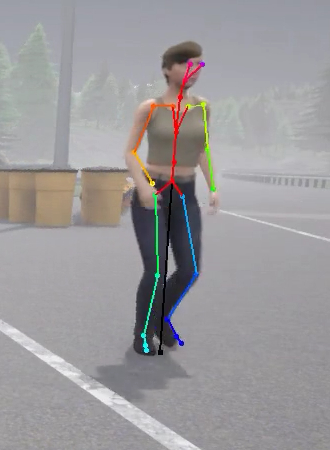}
    \caption{(Left) CARLA skeleton in T-pose with joints' names and indices annotated. (Right) A zoomed-in look at the overlayed skeleton as it appears in one of the dataset frames. \label{fig:concepts:carla:skeleton}}
\end{figure}

In our work, we use a subset of the full CARLA skeleton that consists of 26 joints - 25 joints that correspond to what can be thought of as a ,,physical'' human skeleton and one meta point, the root point, that is used as a reference for other points. We disregard the joints representing human hands as the resolution of images we expect to work with at any point would be insufficient to distinguish them. When describing a skeleton, a concept of T-pose needs to be mentioned. T-pose is a basic position the model is developed with. It is also the default position a model will appear in when added to the simulation for the first time (spawned). As can be guessed, the name comes from a position where a human appears like a ,,T'' letter. The identifiers of the joints and a close-up of the skeleton as it appears in the example frame can be seen in Fig.~\ref{fig:concepts:carla:skeleton}.

Actors in CARLA can be understood as entities we can interact with. It may be a pedestrian, camera, vehicle, or other object. Actors are spawned from blueprints -- templates that hold object attributes, some configurable, and have associated information on how the object should behave and look. The available blueprints can be read from the library. We are particularly interested in pedestrian blueprints, which start with \texttt{walker.pedestrian.*}, and sensor blueprints \texttt{sensor.*.*}.

Each \texttt{walker.pedestrian.*} blueprint has some read-only attributes, like age and gender, and some modifiable ones, like speed. As mentioned, pedestrians use four basic skeletons as a base, one for an adult female, an adult male, a child female, and a child male. Each blueprint differs visually in clothing choice, hairstyle, and skin color. Included are some professional outfits, e.g. a policeman, but as they may differ between the countries, they should not be used to differentiate between civilians and uniformed pedestrians.

The sensors in CARLA that can be created and used to receive data about the world can be divided into several groups. Of most interest in our case are various cameras, with sensors belonging to the LIDAR class planned for addition soon. Sensor blueprints offer a significantly wider selection of modifiable attributes than \texttt{pedestrian.walker.*} ones, allowing a high degree of customization.

All \texttt{sensor.camera.*} blueprints share some common characteristics. Among the most useful is the possibility to modify the size of the image returned by the camera and its field of vision (FOV). Currently, we capture the data from the RGB camera and semantic segmentation camera. Dynamic Vision Sensor (DVS), also known as Events camera, is in the testing phase. 

\subsection{Overall architecture}
\label{ssec:concepts:architecture}

We believe that the easiest way to utilize CARLA, especially on a remote server, is by using the Docker image provided by a CARLA team. Currently, we use the image that provides version 0.9.13. Contrary to what seems a common approach, we do not add our code to that base image, as that has the potential to generate many conflicts, considering our use cases. Instead, we take advantage of the possibilities offered by Docker Compose to connect several containers together.

Describing the philosophy and usage of containers and related tools is beyond the scope of this article, and to understand this work, it is enough to grasp its similarities to (probably more common) virtual machines. To create a container, an image is needed. The image acts as a template, similar to how a class definition is a template to create instances in object programming. Several containers can run on one physical machine simultaneously and be connected via a private network. They also can share volumes used to store data. One of the crucial advantages of Docker containers is an option to specify \texttt{Dockerfile}, in which the contents of the image can be specified using plain text. It is also possible to extend some base images with one's specific needs. We invite the reader to explore \href{https://docs.docker.com/get-started/overview/}{the Docker documentation website} for more information about containers and Docker.

\begin{figure}
    \centering
    \includegraphics[width=0.9\textwidth]{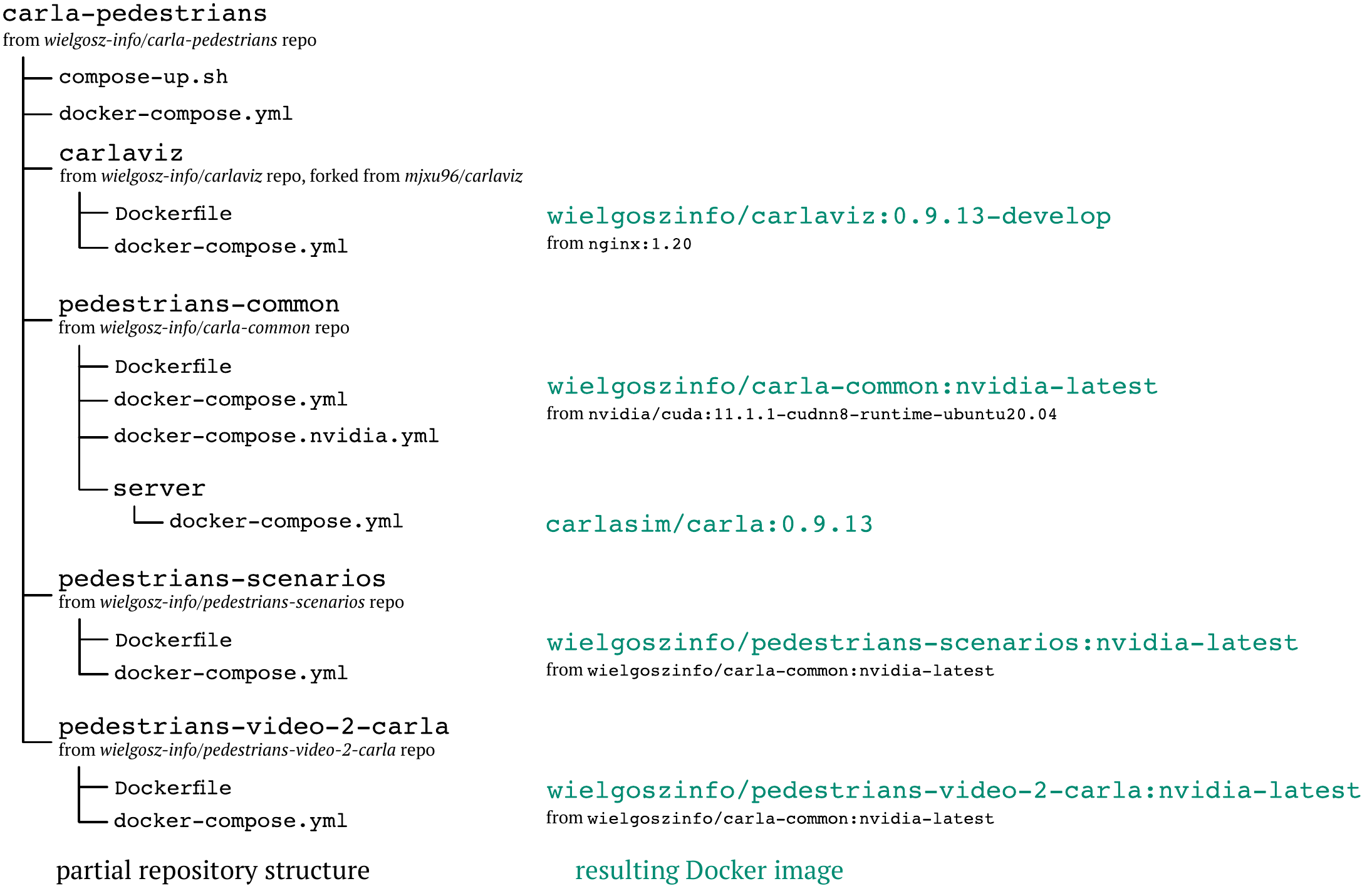}
    \caption{Partial structure of the main repository and its relation to Docker images. All mentioned repositories are hosted on GitHub. \label{fig:concepts:architecture:main_repo}}
\end{figure}

In our main code repository, \texttt{carla-pedestrians}, we specify how the containers we use should work together. The specific image contents are defined either in subdirectories or in Git submodules, using \texttt{docker-compose.yml} and \texttt{Dockerfile} files. A diagram of the high-level container structure can be found in Fig.~\ref{fig:concepts:architecture:main_repo}. Not all containers need to be running simultaneously, but in general \texttt{pedestrians-scenarios} requires the CARLA server to be running; \texttt{pedestrians-video-2-carla} only needs it if outputs rendering in CARLA is requested. CarlaViz \cite{xu2019carlaviz} allows an in-browser preview of what is happening on the server, but it requires the hero vehicle to be present; it is not useful during dataset generation and was included only for completeness.

\subsection{Dataset generation}
\label{ssec:concepts:generation}

The \texttt{pedestrians-scenarios} repository and corresponding container allow to, among others, record and save data from CARLA to create a dataset. The code is organized into several logical modules, but two are important for dataset generation. \texttt{karma} module contains various lower-level definitions and logic that potentially can be reused in other applications. \texttt{datasets} module provides the actual generation flow, with customization done mostly at the level of BatchGenerators. Some of the simpler, non-behavioral modifications are also possible with the configuration files, examples of which can be found in \texttt{pedestrians-scenarios/configs}.

The following list contains some concepts we use concerning dataset generation.

\begin{itemize}
    \item \textbf{World} -- whenever we write about the world, we mean a newly loaded carla.World instance with a specific carla.Map.
    \item \textbf{Scenario} -- a template used to create scenes. It describes the number of actors and their possible behaviors. It may be as simple as specifying that there should be a single pedestrian that can either walk along the street or cross it, and there is a \SI{50}{\percent} chance for each. Generally, an arbitrary number of pedestrians can participate in a single scenario, although we have only implemented single-pedestrian scenarios.
    \item \textbf{Scene} or \textbf{clip} -- an randomized instance of the specific scenario that takes place inside the world. Pedestrians (and sensors) exist in the world for the whole duration of the scene.
    \item \textbf{Batch} -- all scenes simulated synchronously. The generator can simultaneously capture several clips in one world to save computation time. However, since the starting points are randomized, actors belonging to the parallel scenes may get captured on the sensors; therefore, this option should be used carefully. By default, the batch size is set to one, so each clip will be recorded in a newly loaded world. The batch size can be modified via the config file.
    \item \textbf{Pedestrian profile} -- a structure describing possible types of pedestrians. For now, it includes fields like age, gender, walking speed (expressed as mean and standard deviation), and crossing speed (likewise). We prepared four basic profiles, one for each available age-gender combination. It is possible to modify/add new profiles or change the proportions with which they are used by writing a custom config file. One of the examples, \texttt{pedestrians-scenarios/configs/children.yaml} shows how to create a dataset that will contain only children, with a higher probability of females.
    \item \textbf{Camera} -- this is a slightly misleading term for what could be more aptly named ,,point of view''. Several sensors can be associated with each camera (actual RGB, DVS or semantic cameras, LIDAR, etc.). They are, however, synchronized in their physical orientation in space and, where possible, attributes like returned image size. This results in synchronized data, creating an illusion of a single device operating in multiple ,,modes'' at the same time. The \textbf{camera.idx} used later in the dataset description refers to such a logical device. We plan to add the ability to specify the required modes via a config file in the future. It is already possible to add several points of view, using \texttt{camera\_position\_distributions} field.
\end{itemize}

One of the available generators is BinarySinglePedestrian, which was used to generate the dataset described in this work. It uses default pedestrian profiles, each having a \SI{25}{\percent} chance to be used in a scenario. It also uses only crossing speed, and the pedestrian's target speed does not change during the scene. The actual speed can be different when forced by the obstacle. A single default camera is used to capture the scene, and its position is sampled from the following distributions: x-axis mean=-7.0, std=2.0 (about seven meters in front of the scene); y-axis mean=0.0, std=0.25 (looking mostly in the perpendicular direction), z-axis mean=1.0, std=0.25 (about one meter above the ground).

\section{Binary Single Pedestrian dataset}
\label{sec:dataset}

\begin{table}[btp]
    \small
    \begin{tabularx}{\textwidth}{@{}llllX@{}}
       \toprule
       \textbf{pedestrian.age} & \textbf{pedestrian.gender} & \textbf{no. of scenes} & \textbf{no. of frames} & \textbf{frame.pedestrian.is\_crossing} \\
       \midrule
       \multirow{2}{*}{adult} & female & \num{88}  & \num{49951}  & \num{17820} \\
                              & male   & \num{109} & \num{60775}  & \num{22924} \\
       \multirow{2}{*}{child} & female & \num{100} & \num{55755}  & \num{23038} \\
                              & male   & \num{99}  & \num{58846}  & \num{28082} \\
       \midrule
                              &        & \num{396} & \num{225327} & \num{91864} \\
       \bottomrule
    \end{tabularx}
    \caption{Basic dataset stats including the number of annotated frames. \label{tab:dataset:stats}}
\end{table}

The dataset contains almost 400 scenes. Each clip was recorded for \num{900} frames at \SI{30}{\fps}, yielding \SI{30}{\second} of video and associated semantic segmentation data. We refer to it as binary, as it contains only two classes for use in classification tasks: the pedestrian is either crossing or non-crossing. The detailed stats can be found in Tab.~\ref{tab:dataset:stats}.

As mentioned previously, each video contains only a single pedestrian. It may contain only a fragment of a silhouette, and there is no guarantee that the pedestrian is visible in all frames (often, there are only background frames at the beginning and end of the video). Each frame where the pedestrian is in the viewport has a corresponding entry in the annotations file (\texttt{data.csv}). \SI{63.22}{\percent} of frames are annotated.

The annotations are provided in CSV format. It was decided that the ease of working with and previewing such data trumps its bigger size. When running actual tasks, this data is split into clips of the desired length and saved in efficient HDF5 format. The annotations consist of 24 columns, including 20 data columns and four used for indexing purposes. An example of a single data row can be found in Tab.~\ref{tab:dataset:example_data_row}. The index columns are:

\begin{enumerate}
    \item \textbf{id} -- unique ID of the recorded scene/clip,
    \item \textbf{camera.idx} -- index of the camera/point of view used to capture the data,
    \item \textbf{frame.idx} -- frame index in the corresponding video and other files that this row applies to; in general \textbf{it is not continuous} as only frames in which the pedestrian is visible are annotated, while accompanying source files (videos, APNGs with semantic labels) contain all frames as recorded,
    \item \textbf{pedestrian.idx} -- the pedestrian the data in the row corresponds to; it is unique only in the scene. It is always zero, as all scenarios have only one pedestrian. In the future, we plan on introducing scenarios that involve several actors at the same time.
\end{enumerate}

Data columns can be split into two main groups. First of them are initial attributes that are constant for the whole clip:

\begin{enumerate}[resume]
    \item \textbf{world.map} -- this is the name of the CARLA map that was used to record the scene,
    \item \textbf{camera.transform} -- world coordinates of the camera with index \textbf{camera.idx}. Currently, cameras are assumed to be static; in the future, we plan to add moving cameras (like on vehicles),
    \item \textbf{camera.width} -- pixel width of the image captured by the camera \textbf{camera.idx},
    \item \textbf{camera.height} -- like above, but for height,
    \item \textbf{camera.recording} -- the relative path to the \texttt{*.mp4} file that contains the RGB recording of the \textbf{camera.idx} camera,
    \item \textbf{camera.semantic\_segmentation} -- the relative path to the \texttt{*.apng} file that contains the semantic segmentation labels of the \textbf{camera.idx} camera,
    \item \textbf{pedestrian.model} -- name of the blueprint (,,skin'') that was used to create the pedestrian with index \textbf{pedestrian.idx},
    \item \textbf{pedestrian.age} -- age category as specified in blueprint; currently can be either \textit{adult} or \textit{child},
    \item \textbf{pedestrian.gender} -- gender category as specified in blueprint; currently can be either \textit{female} or \textit{male},
    \item \textbf{pedestrian.spawn\_point} -- world coordinates where the pedestrian with index \textbf{pedestrian.idx} first appeared.
\end{enumerate}

\noindent
The second group contains values that can change in each frame, and most are specific to a particular pedestrian:

\begin{enumerate}[resume]
    \item \textbf{world.frame} -- which simulation frame is this since the world was first created; there can be a scenario-dependent number of ,,startup frames'' needed before the actual scene recording can start,
    \item \textbf{frame.pedestrian.is\_crossing} -- is the pedestrian considered to be crossing the street in this frame; this is an automatic annotation that checks whether the pedestrian is on a driving lane,
    \item \textbf{frame.pedestrian.transform} -- pedestrian's bounding box center coordinates in the world,
    \item \textbf{frame.pedestrian.velocity} -- pedestrian's velocity in \si{\meter \per \second},
    \item \textbf{frame.pedestrian.pose.in\_frame} -- is at least one pedestrian's joint in the camera viewport; it does not mean it is actually visible, as the obstacles may obscure the view, only that its coordinates fall into the correct range,
    \item \textbf{frame.pedestrian.pose.in\_segmentation} -- is at least one pixel in the segmentation mask marked as a pedestrian, i.e. at least a fragment of the pedestrian is in the viewport and any possible cover is less than \SI{100}{\percent},
    \item \textbf{frame.pedestrian.pose.world} -- position of each joint expressed in world coordinates,
    \item \textbf{frame.pedestrian.pose.component} -- position of each joint relative to the pivot of the actor,
    \item \textbf{frame.pedestrian.pose.relative} -- position of each joint relative to its parent in a kinematic tree,
    \item \textbf{frame.pedestrian.pose.camera} -- position of each joint projected to 2D with camera \textbf{camera.idx}.
\end{enumerate}

\begin{table}[ptb]
    \scriptsize
    \sisetup{round-mode=places, round-precision=4}
    \setlength{\tabcolsep}{2.5pt}
    \begin{tabularx}{\textwidth}{@{}Xr*{7}{r}l@{}}
       \toprule
       \multicolumn{2}{@{}l}{\textbf{name}} & \multicolumn{8}{l}{\enspace\textbf{value}} \\
       \midrule
       \multicolumn{2}{@{}l}{id} & \multicolumn{8}{l}{\enspace 094b9fe1-babe-48d5-bd17-3ba3185690c5-0} \\
       \multicolumn{2}{@{}l}{camera.idx} & \multicolumn{8}{l}{\enspace 0} \\
       \multicolumn{2}{@{}l}{frame.idx} & \multicolumn{8}{l}{\enspace 0} \\
       \multicolumn{2}{@{}l}{pedestrian.idx} & \multicolumn{8}{l}{\enspace 0} \\
       \midrule
       \multicolumn{2}{@{}l}{world.map} & \multicolumn{8}{l}{\enspace /Game/Carla/Maps/Town04} \\
       camera.transform & \textit{\tiny[x, y, z (\si{\meter}), pitch, yaw, roll (\si{\deg})]} & [ & \num{-5.622942924499512}, & \num{316.14385986328125}, & \num{1.1189192533493042}, & \num{-9.404664993286133}, & \num{-65.50199890136719}, & \num{0.0000004327028477746353} & ] \\
       camera.width & {\tiny\si{\pixel}} & \multicolumn{8}{l}{\enspace 1600} \\
       camera.height & {\tiny\si{\pixel}} & \multicolumn{8}{l}{\enspace 600} \\
       \multicolumn{2}{@{}l}{camera.recording} & \multicolumn{8}{l}{\enspace\texttt{clips/094b9fe1-babe-48d5-bd17-3ba3185690c5-0-0.mp4}} \\
       \multicolumn{2}{@{}l}{camera.semantic\_segmentation} & \multicolumn{8}{l}{\enspace\texttt{clips/094b9fe1-babe-48d5-bd17-3ba3185690c5-0-1.apng}} \\
       \multicolumn{2}{@{}l}{pedestrian.model} & \multicolumn{8}{l}{\enspace walker.pedestrian.0001} \\
       \multicolumn{2}{@{}l}{pedestrian.age} & \multicolumn{8}{l}{\enspace adult} \\
       \multicolumn{2}{@{}l}{pedestrian.gender} & \multicolumn{8}{l}{\enspace female} \\
       pedestrian.spawn\_point & \textit{\tiny[x, y, z (\si{\meter}), pitch, yaw, roll (\si{\deg})]} & [ & \num{0.3316575288772583}, & \num{308.9236145019531}, & \num{0.4300410747528076}, & \num{0.0}, & \num{0.0}, & \num{0.0} & ] \\
       \midrule
       \multicolumn{2}{@{}l}{world.frame} & \multicolumn{8}{l}{\enspace 9} \\
       \multicolumn{2}{@{}l}{frame.pedestrian.is\_crossing} & \multicolumn{8}{l}{\enspace False} \\
       frame.pedestrian.transform & \textit{\tiny[x, y, z (\si{\meter}), pitch, yaw, roll (\si{\deg})]} & [ & \num{0.33056285977363586}, & \num{308.92596435546875}, & \num{0.9514997601509094}, & \num{0.0}, & \num{114.87537384033203}, & \num{0.0} & ] \\
       frame.pedestrian.velocity & \textit{\tiny[x, y, z (\si{\meter\per\second})]} & [ & \num{-0.03283996134996414}, & \num{0.07089842855930328}, & \num{0.0} & \multicolumn{4}{l}{]} \\
       \multicolumn{2}{@{}l}{frame.pedestrian.pose.in\_frame} & \multicolumn{8}{l}{\enspace True} \\
       \multicolumn{2}{@{}l}{frame.pedestrian.pose.in\_segmentation} & \multicolumn{8}{l}{\enspace True} \\
       frame.pedestrian.pose.world & 
                                   & [[ & \num{0.3283732831478119}, & \num{308.9306945800781}, & \num{0.03149978443980217}, & \num{0.0}, & \num{24.87514305114746}, & \num{89.99618530273438} & ], \\
                                   & \textit{\tiny[[x, y, z (\si{\meter}), pitch, yaw, roll (\si{\deg})], \ldots]}
                                   & [ & \num{0.32617998123168945}, & \num{308.9266357421875}, & \num{1.0833592414855957}, & \num{-0.38330337405204773}, & \num{22.558658599853516}, & \num{91.07976531982422} & ], \\
                                   & & [ & \num{0.33041754364967346}, & \num{308.9183044433594}, & \num{1.1900527477264404}, & \num{0.05510596185922623}, & \num{21.507951736450195}, & \num{91.76720428466797} & ], \\
                                   & & [ & \num{0.3408958315849304}, & \num{308.8912658691406}, & \num{1.3531420230865479}, & \num{0.4752991795539856}, & \num{19.955116271972656}, & \num{91.16964721679688} & ], \\
                                   & & [ & \num{0.3753882050514221}, & \num{308.9134216308594}, & \num{1.5019692182540894}, & \num{-3.699148178100586}, & \num{8.623555183410645}, & \num{95.4970932006836} & ], \\
                                   & & [ & \num{0.48325881361961365}, & \num{308.92974853515625}, & \num{1.49491548538208}, & \num{-77.19293212890625}, & \num{22.78913688659668}, & \num{-80.14524841308594} & ], \\
                                   & & [ & \num{0.5425371527671814}, & \num{308.95562744140625}, & \num{1.2331812381744385}, & \num{-73.68435668945312}, & \num{89.91813659667969}, & \num{-139.34934997558594} & ], \\
                                   & & [ & \num{0.5426682233810425}, & \num{309.02557373046875}, & \num{0.9940587282180786}, & \num{-2.686436176300049}, & \num{-53.12208938598633}, & \num{167.73931884765625} & ], \\
                                   & & [ & \num{0.3419244587421417}, & \num{308.8836975097656}, & \num{1.5485349893569946}, & \num{0.8279008269309998}, & \num{-157.03587341308594}, & \num{77.15479278564453} & ], \\
                                   & & [ & \num{0.331692636013031}, & \num{308.9111328125}, & \num{1.6377370357513428}, & \num{2.2554445266723633}, & \num{-154.06863403320312}, & \num{74.4913330078125} & ], \\
                                   & & [ & \num{0.32486429810523987}, & \num{309.0069580078125}, & \num{1.710412859916687}, & \num{-2.26051926612854}, & \num{25.957834243774414}, & \num{96.48550415039062} & ], \\
                                   & & [ & \num{0.26577135920524597}, & \num{308.9782409667969}, & \num{1.713000774383545}, & \num{-2.26051926612854}, & \num{25.957834243774414}, & \num{96.48550415039062} & ], \\
                                   & & [ & \num{0.29791924357414246}, & \num{308.8852844238281}, & \num{1.5012900829315186}, & \num{-1.5654792785644531}, & \num{-152.72189331054688}, & \num{-93.72132873535156} & ], \\
                                   & & [ & \num{0.20078250765800476}, & \num{308.8351745605469}, & \num{1.4983031749725342}, & \num{76.90486907958984}, & \num{16.13275146484375}, & \num{105.75209045410156} & ], \\
                                   & & [ & \num{0.13712742924690247}, & \num{308.8182067871094}, & \num{1.2368695735931396}, & \num{72.41229248046875}, & \num{-65.76908111572266}, & \num{30.698078155517578} & ], \\
                                   & & [ & \num{0.10640466213226318}, & \num{308.8881530761719}, & \num{0.9997186064720154}, & \num{2.2516672611236572}, & \num{80.10244750976562}, & \num{-25.211469650268555} & ], \\
                                   & & [ & \num{0.2587330937385559}, & \num{308.88128662109375}, & \num{0.9965510964393616}, & \num{5.23007345199585}, & \num{40.1851921081543}, & \num{-97.50977325439453} & ], \\
                                   & & [ & \num{0.24485188722610474}, & \num{308.93212890625}, & \num{0.5368066430091858}, & \num{2.7538979053497314}, & \num{40.12872314453125}, & \num{98.49413299560547} & ], \\
                                   & & [ & \num{0.27762117981910706}, & \num{308.8830871582031}, & \num{0.10569610446691513}, & \num{0.48394620418548584}, & \num{43.34621810913086}, & \num{-159.99436950683594} & ], \\
                                   & & [ & \num{0.20286335051059723}, & \num{308.9625549316406}, & \num{0.07081379741430283}, & \num{-4.248117923736572}, & \num{134.85377502441406}, & \num{90.53681945800781} & ], \\
                                   & & [ & \num{0.17112523317337036}, & \num{308.9945983886719}, & \num{0.05563376471400261}, & \num{-0.7624198198318481}, & \num{-135.1571807861328}, & \num{-166.28302001953125} & ], \\
                                   & & [ & \num{0.40481075644493103}, & \num{308.9419860839844}, & \num{0.9954928159713745}, & \num{-4.093700885772705}, & \num{11.055594444274902}, & \num{85.5928955078125} & ], \\
                                   & & [ & \num{0.387456476688385}, & \num{308.96185302734375}, & \num{0.5334904789924622}, & \num{-0.7111387848854065}, & \num{10.852561950683594}, & \num{-77.36528015136719} & ], \\
                                   & & [ & \num{0.4259209632873535}, & \num{308.8780517578125}, & \num{0.10824508219957352}, & \num{1.4687979221343994}, & \num{6.067918300628662}, & \num{-156.92257690429688} & ], \\
                                   & & [ & \num{0.41574957966804504}, & \num{308.9846496582031}, & \num{0.06758200377225876}, & \num{-5.178396224975586}, & \num{94.5060806274414}, & \num{91.12165832519531} & ], \\
                                   & & [ & \num{0.41246160864830017}, & \num{309.0293884277344}, & \num{0.05167381092905998}, & \num{-0.8877537846565247}, & \num{-175.35214233398438}, & \num{12.791315078735352} & ]]
    \end{tabularx}
    \caption{Example data from a single row in \texttt{data.csv}. Coordinates were rounded to four decimal places. Table spans several pages.\label{tab:dataset:example_data_row}}
\end{table}
\begin{table}[ptb]
    \ContinuedFloat
    \scriptsize
    \sisetup{round-mode=places, round-precision=4}
    \setlength{\tabcolsep}{2.5pt}
    \begin{tabularx}{\textwidth}{@{}Xl*{7}{r}l@{}}
       frame.pedestrian.pose.component &
                                   & [[ & \num{0.0}, & \num{0.0}, & \num{-0.9199999570846558}, & \num{0.0}, & \num{-90.00020599365234}, & \num{89.99618530273438} & ], \\
                                   & \textit{\tiny[[x, y, z (\si{\meter}), pitch, yaw, roll (\si{\deg})], \ldots]}
                                   & [ & \num{-0.0027629979886114597}, & \num{0.0036986898630857468}, & \num{0.1318594366312027}, & \num{-0.38330337405204773}, & \num{-92.31671142578125}, & \num{91.07976531982422} & ], \\
                                   & & [ & \num{-0.012111622840166092}, & \num{0.003362396964803338}, & \num{0.2385530024766922}, & \num{0.05510596185922623}, & \num{-93.367431640625}, & \num{91.76720428466797} & ], \\
                                   & & [ & \num{-0.041042692959308624}, & \num{0.005226826760917902}, & \num{0.4016422927379608}, & \num{0.4752991795539856}, & \num{-94.92027282714844}, & \num{91.16964721679688} & ], \\
                                   & & [ & \num{-0.03545817360281944}, & \num{-0.035382166504859924}, & \num{0.5504695177078247}, & \num{-3.699148178100586}, & \num{-106.25186157226562}, & \num{95.4970932006836} & ], \\
                                   & & [ & \num{-0.06600254029035568}, & \num{-0.14012162387371063}, & \num{0.5434156656265259}, & \num{-77.19293212890625}, & \num{-92.08619689941406}, & \num{-80.14525604248047} & ], \\
                                   & & [ & \num{-0.06747747212648392}, & \num{-0.2047780603170395}, & \num{0.28168150782585144}, & \num{-73.68437957763672}, & \num{-24.957246780395508}, & \num{-139.34934997558594} & ], \\
                                   & & [ & \num{-0.0040623280219733715}, & \num{-0.23432578146457672}, & \num{0.042558975517749786}, & \num{-2.686436176300049}, & \num{-167.9974365234375}, & \num{167.73931884765625} & ], \\
                                   & & [ & \num{-0.04835044592618942}, & \num{0.007481341250240803}, & \num{0.5970352292060852}, & \num{0.8279076814651489}, & \num{88.08875274658203}, & \num{77.15479278564453} & ], \\
                                   & & [ & \num{-0.019133226945996284}, & \num{0.005212557036429644}, & \num{0.6862373352050781}, & \num{2.255451202392578}, & \num{91.05598449707031}, & \num{74.4913330078125} & ], \\
                                   & & [ & \num{0.07066965103149414}, & \num{-0.028899049386382103}, & \num{0.7589130997657776}, & \num{-2.26051926612854}, & \num{-88.91751861572266}, & \num{96.48550415039062} & ], \\
                                   & & [ & \num{0.0694618821144104}, & \num{0.03679689019918442}, & \num{0.7615010142326355}, & \num{-2.26051926612854}, & \num{-88.91751861572266}, & \num{96.48550415039062} & ], \\
                                   & & [ & \num{-0.028386835008859634}, & \num{0.046730272471904755}, & \num{0.5497903227806091}, & \num{-1.5654724836349487}, & \num{92.40272521972656}, & \num{-93.72132873535156} & ], \\
                                   & & [ & \num{-0.03297647461295128}, & \num{0.15592852234840393}, & \num{0.5468034148216248}, & \num{76.9048843383789}, & \num{-98.74272918701172}, & \num{105.75212097167969} & ], \\
                                   & & [ & \num{-0.021598204970359802}, & \num{0.22081749141216278}, & \num{0.285369873046875}, & \num{72.41229248046875}, & \num{179.35556030273438}, & \num{30.698091506958008} & ], \\
                                   & & [ & \num{0.05477777495980263}, & \num{0.21926933526992798}, & \num{0.04821883887052536}, & \num{2.2516672611236572}, & \num{-34.7729606628418}, & \num{-25.211469650268555} & ], \\
                                   & & [ & \num{-0.015518064610660076}, & \num{0.08395694196224213}, & \num{0.045051343739032745}, & \num{5.23007345199585}, & \num{-74.6901626586914}, & \num{-97.50977325439453} & ], \\
                                   & & [ & \num{0.03644418716430664}, & \num{0.07516472041606903}, & \num{-0.41469308733940125}, & \num{2.7538979053497314}, & \num{-74.74665069580078}, & \num{98.49413299560547} & ], \\
                                   & & [ & \num{-0.02185080014169216}, & \num{0.06607356667518616}, & \num{-0.8458036780357361}, & \num{0.48394620418548584}, & \num{-71.52915954589844}, & \num{-159.99436950683594} & ], \\
                                   & & [ & \num{0.08169534802436829}, & \num{0.10046584904193878}, & \num{-0.8806859254837036}, & \num{-4.248117923736572}, & \num{19.97840118408203}, & \num{90.53681182861328} & ], \\
                                   & & [ & \num{0.12412313371896744}, & \num{0.1157774105668068}, & \num{-0.8958659768104553}, & \num{-0.7624198198318481}, & \num{109.96742248535156}, & \num{-166.28302001953125} & ], \\
                                   & & [ & \num{-0.021911410614848137}, & \num{-0.07409465312957764}, & \num{0.04399307072162628}, & \num{-4.093700885772705}, & \num{-103.81983184814453}, & \num{85.59290313720703} & ], \\
                                   & & [ & \num{0.003426780691370368}, & \num{-0.06671404838562012}, & \num{-0.41800928115844727}, & \num{-0.7111455798149109}, & \num{-104.02286529541016}, & \num{-77.36527252197266} & ], \\
                                   & & [ & \num{-0.08878640085458755}, & \num{-0.06635623425245285}, & \num{-0.8432546854019165}, & \num{1.4688048362731934}, & \num{-108.80748748779297}, & \num{-156.92257690429688} & ], \\
                                   & & [ & \num{0.012203731574118137}, & \num{-0.10197003930807114}, & \num{-0.8839177489280701}, & \num{-5.178396224975586}, & \num{-20.369295120239258}, & \num{91.12165832519531} & ], \\
                                   & & [ & \num{0.0541638545691967}, & \num{-0.11780117452144623}, & \num{-0.899825930595398}, & \num{-0.8877537846565247}, & \num{69.7724609375}, & \num{12.791315078735352} & ]] \\
       frame.pedestrian.pose.relative &
                                   & [[ & \num{0.0}, & \num{0.0}, & \num{0.0}, & \num{0.0}, & \num{0.0000012074581263732398}, & \num{89.99617767333984} & ], \\
                                   & \textit{\tiny[[x, y, z (\si{\meter}), pitch, yaw, roll (\si{\deg})], \ldots]}
                                   & [ & \num{-0.0036986921913921833}, & \num{-1.0518593788146973}, & \num{-0.0026930999010801315}, & \num{-2.316465377807617}, & \num{0.3834635019302368}, & \num{1.0680601596832275} & ], \\
                                   & & [ & \num{0.000000013589858838258806}, & \num{-0.10650123655796051}, & \num{-0.01133633591234684}, & \num{-1.058788537979126}, & \num{-0.4185433089733124}, & \num{0.6883028149604797} & ], \\
                                   & & [ & \num{-0.000004966258984495653}, & \num{-0.16211774945259094}, & \num{-0.03400629758834839}, & \num{-1.5650148391723633}, & \num{-0.3722708523273468}, & \num{-0.5852881073951721} & ], \\
                                   & & [ & \num{0.04121358320116997}, & \num{-0.14864413440227509}, & \num{0.006013965699821711}, & \num{-11.218653678894043}, & \num{4.4800543785095215}, & \num{3.5671911239624023} & ], \\
                                   & & [ & \num{0.10933008044958115}, & \num{0.0000012266635849300656}, & \num{-0.000011384188837837428}, & \num{8.388673782348633}, & \num{73.71633911132812}, & \num{-157.85984802246094} & ], \\
                                   & & [ & \num{0.2695574462413788}, & \num{0.005116452928632498}, & \num{-0.000004158020146860508}, & \num{-15.855490684509277}, & \num{-3.600116014480591}, & \num{6.7384514808654785} & ], \\
                                   & & [ & \num{0.24914677441120148}, & \num{0.000002248287955808337}, & \num{0.00004514693500823341}, & \num{79.2442855834961}, & \num{-163.8402099609375}, & \num{14.058793067932129} & ], \\
                                   & & [ & \num{0.0000015163420812314143}, & \num{-0.19520632922649384}, & \num{-0.011461276561021805}, & \num{-3.034700632095337}, & \num{-178.75738525390625}, & \num{168.28225708007812} & ], \\
                                   & & [ & \num{-0.000004050731604365865}, & \num{-0.09348762035369873}, & \num{-0.008710688911378384}, & \num{3.2087066173553467}, & \num{-0.7350655198097229}, & \num{-2.7638869285583496} & ], \\
                                   & & [ & \num{-0.03287297859787941}, & \num{-0.09517118334770203}, & \num{-0.06611914187669754}, & \num{-0.026842642575502396}, & \num{-179.997802734375}, & \num{170.9779052734375} & ], \\
                                   & & [ & \num{0.03288498893380165}, & \num{-0.09517201036214828}, & \num{-0.06612218916416168}, & \num{-0.026842642575502396}, & \num{-179.997802734375}, & \num{170.9779052734375} & ], \\
                                   & & [ & \num{-0.04120556637644768}, & \num{-0.14864873886108398}, & \num{0.006016416475176811}, & \num{-7.296198844909668}, & \num{178.74696350097656}, & \num{-2.3411991596221924} & ], \\
                                   & & [ & \num{0.10933545231819153}, & \num{-0.000000424385063979571}, & \num{-0.000007634163011971395}, & \num{-6.113681316375732}, & \num{104.4920425415039}, & \num{25.683034896850586} & ], \\
                                   & & [ & \num{-0.2695578634738922}, & \num{-0.005109663121402264}, & \num{-0.000004138946223974926}, & \num{-19.590484619140625}, & \num{-5.28346586227417}, & \num{5.751477241516113} & ], \\
                                   & & [ & \num{-0.24914763867855072}, & \num{0.0000007247925850606407}, & \num{0.0013483624206855893}, & \num{77.00750732421875}, & \num{160.926513671875}, & \num{-34.75770950317383} & ], \\
                                   & & [ & \num{-0.07909451425075531}, & \num{0.0876249447464943}, & \num{-0.014340247958898544}, & \num{17.437143325805664}, & \num{-6.207947254180908}, & \num{169.70538330078125} & ], \\
                                   & & [ & \num{-0.019800415262579918}, & \num{-0.4621564447879791}, & \num{0.012713927775621414}, & \num{0.37944430112838745}, & \num{-2.447601795196533}, & \num{-164.00027465820312} & ], \\
                                   & & [ & \num{-0.027271423488855362}, & \num{0.43423792719841003}, & \num{0.005568313412368298}, & \num{3.5171782970428467}, & \num{1.7688037157058716}, & \num{101.47488403320312} & ], \\
                                   & & [ & \num{-0.00010953425953630358}, & \num{-0.11445825546979904}, & \num{-0.004549493547528982}, & \num{-15.763010025024414}, & \num{-91.59938049316406}, & \num{-89.39966583251953} & ], \\
                                   & & [ & \num{0.04610651358962059}, & \num{0.011797547340393066}, & \num{0.0000045394895096251275}, & \num{89.76425170898438}, & \num{73.15614318847656}, & \num{-178.8762664794922} & ], \\
                                   & & [ & \num{0.07908985763788223}, & \num{0.08762496709823608}, & \num{-0.014339422807097435}, & \num{-11.39985179901123}, & \num{4.011956691741943}, & \num{-6.33853006362915} & ], \\
                                   & & [ & \num{0.019795626401901245}, & \num{0.4621571898460388}, & \num{-0.01270927395671606}, & \num{0.05735309422016144}, & \num{-3.3881607055664062}, & \num{-162.9683837890625} & ], \\
                                   & & [ & \num{0.027273092418909073}, & \num{-0.43423762917518616}, & \num{-0.005565528757870197}, & \num{5.144894123077393}, & \num{1.0836515426635742}, & \num{-79.47691345214844} & ], \\
                                   & & [ & \num{0.00011146366159664467}, & \num{-0.11445584148168564}, & \num{-0.0045484923757612705}, & \num{-17.853416442871094}, & \num{-88.50569152832031}, & \num{-90.9408187866211} & ], \\
                                   & & [ & \num{0.04609940946102142}, & \num{0.011797904968261719}, & \num{-0.0000051957367759314366}, & \num{89.76592254638672}, & \num{-105.19662475585938}, & \num{-177.2250213623047} & ]]
    \end{tabularx}
    \caption{Example data from a single row in \texttt{data.csv} (continued).}
\end{table}
\begin{table}[tbp]
    \ContinuedFloat
    \scriptsize
    \sisetup{round-mode=places, round-precision=4}
    \setlength{\tabcolsep}{2.5pt}
    \begin{tabularx}{\textwidth}{@{}Xr*{3}{r}ll*{3}{r}l}
       frame.pedestrian.pose.camera & \textit{\tiny[[x, y (\si{\pixel})], [x, y (\si{\pixel})], \ldots]}
                                   & [[ & \num{1013.4298095703125}, & \num{264.52288818359375} & ],
                                   & & [ & \num{1017.1472778320312}, & \num{170.72708129882812} & ], \\
                                   & & [ & \num{1017.3836669921875}, & \num{161.0204620361328} & ],
                                   & & [ & \num{1017.1865234375}, & \num{146.1831512451172} & ], \\
                                   & & [ & \num{1021.576904296875}, & \num{132.52398681640625} & ],
                                   & & [ & \num{1030.2337646484375}, & \num{133.285888671875} & ], \\
                                   & & [ & \num{1034.8846435546875}, & \num{157.14923095703125} & ],
                                   & & [ & \num{1038.1151123046875}, & \num{178.828125} & ], \\
                                   & & [ & \num{1017.5922241210938}, & \num{128.29660034179688} & ],
                                   & & [ & \num{1018.8466186523438}, & \num{119.92573547363281} & ], \\
                                   & & [ & \num{1024.4339599609375}, & \num{112.6378402709961} & ],
                                   & & [ & \num{1018.3949584960938}, & \num{112.40528869628906} & ], \\
                                   & & [ & \num{1014.329345703125}, & \num{132.56112670898438} & ],
                                   & & [ & \num{1004.357666015625}, & \num{132.85638427734375} & ], \\
                                   & & [ & \num{997.7958374023438}, & \num{156.76368713378906} & ],
                                   & & [ & \num{998.7059326171875}, & \num{178.3826446533203} & ], \\
                                   & & [ & \num{1009.314453125}, & \num{178.58578491210938} & ],
                                   & & [ & \num{1009.5126953125}, & \num{220.1184844970703} & ], \\
                                   & & [ & \num{1007.4016723632812}, & \num{257.8045349121094} & ],
                                   & & [ & \num{1006.50146484375}, & \num{261.91229248046875} & ], \\
                                   & & [ & \num{1006.0143432617188}, & \num{263.6994934082031} & ],
                                   & & [ & \num{1023.3164672851562}, & \num{178.67491149902344} & ], \\
                                   & & [ & \num{1021.3909301757812}, & \num{220.22900390625} & ],
                                   & & [ & \num{1017.5359497070312}, & \num{256.9405212402344} & ], \\
                                   & & [ & \num{1022.8543701171875}, & \num{261.49560546875} & ],
                                   & & [ & \num{1025.1929931640625}, & \num{263.3291320800781} & ]] \\
       \bottomrule
    \end{tabularx}
    \caption{Example data from a single row in \texttt{data.csv} (continued).}
\end{table}

\begin{figure}
    \includegraphics[width=\textwidth]{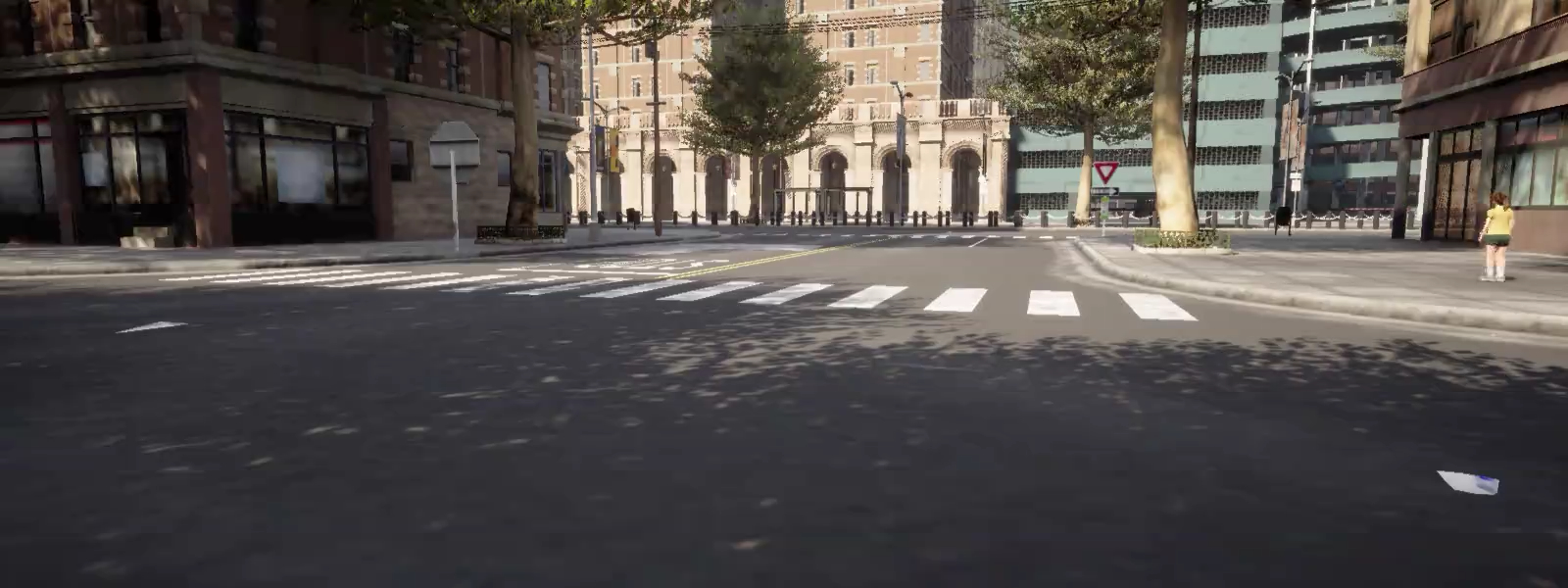}
    \caption{Example frame captured from RGB camera in Town10HD. Visible in the frame is a female child (blueprint walker.pedestrian.0011). \label{fig:dataset:child_female_rgb}}
\end{figure}

\begin{figure}
    \includegraphics[width=\textwidth]{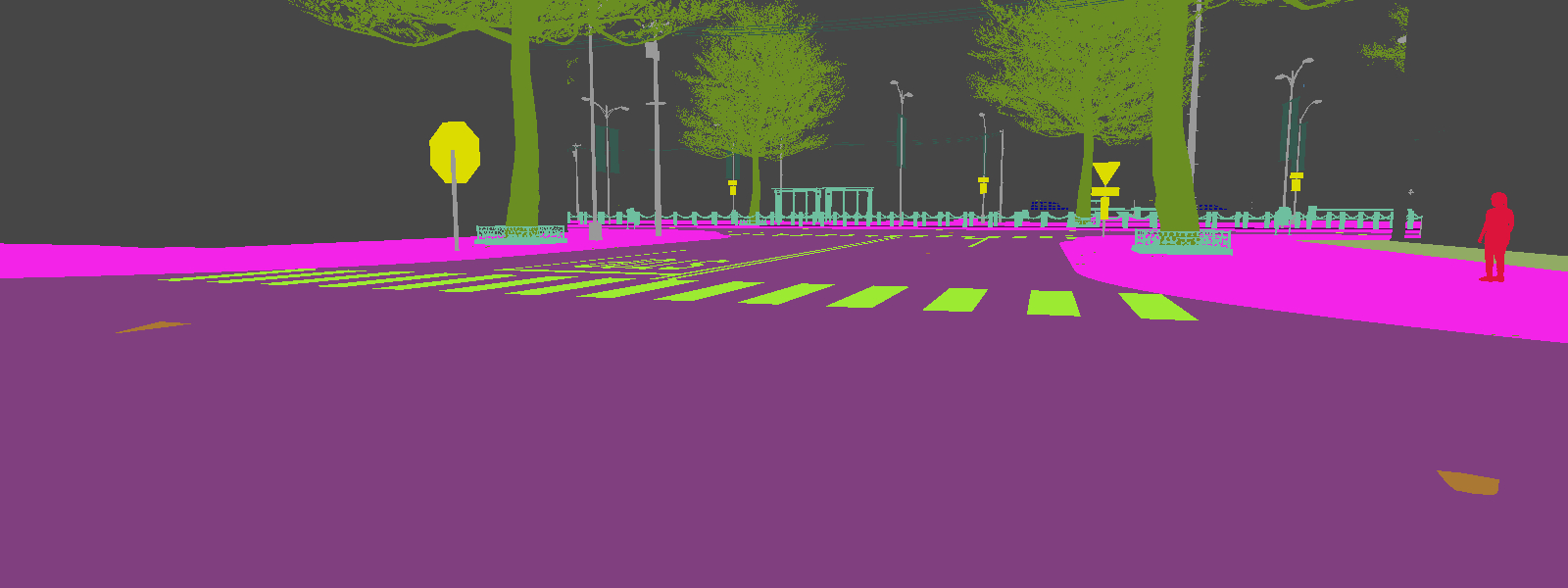}
    \caption{The same frame as in Fig.~\ref{fig:dataset:child_female_rgb}, but captured by the semantic segmentation camera and rendered using the Cityscapes palette. \label{fig:dataset:child_female_segmentation}}
\end{figure}

\begin{figure}
    \includegraphics[width=\textwidth]{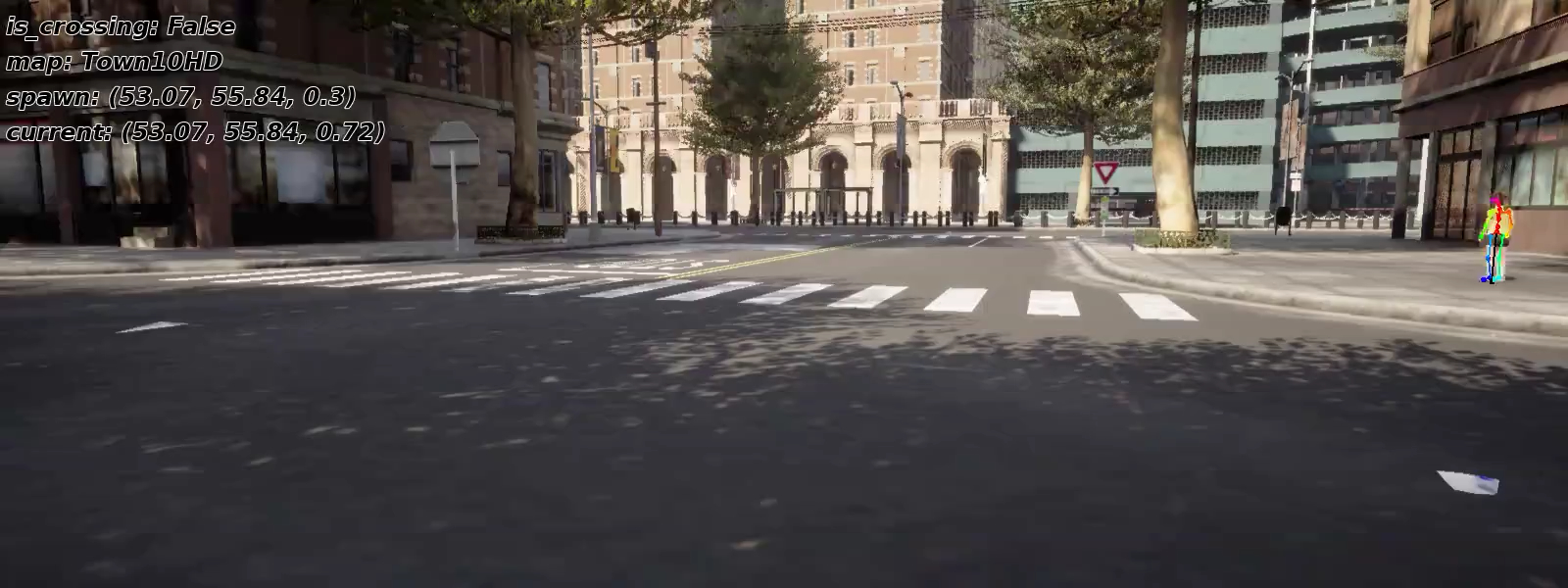}
    \includegraphics[width=\textwidth]{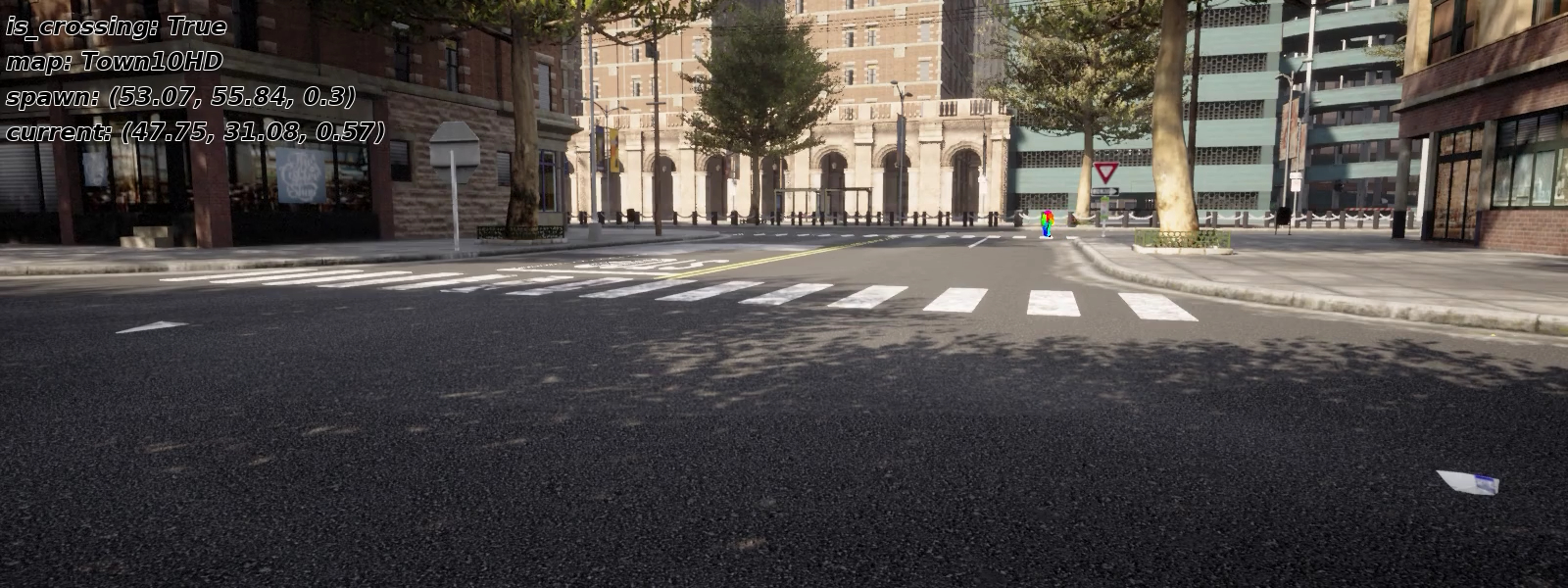}
    \caption{(Above) The same frame as in Fig.~\ref{fig:dataset:child_female_rgb}, but with additional info (including skeleton overlay) rendered using \texttt{dataset\_advanced\_preview.ipynb} notebook from \texttt{pedestrian\_scenarios}. (Below) Different frame from the same video. \label{fig:dataset:child_female_info}}
\end{figure}

The example of captured ground truth data can be seen in Fig.~\ref{fig:dataset:child_female_rgb}, \ref{fig:dataset:child_female_segmentation} and \ref{fig:dataset:child_female_info}.

\section{Example use cases}
\label{sec:examples}

Our aim for creating and sharing this dataset is to encourage feedback. We would also like to facilitate the comparison with other datasets in various possible use cases. Therefore, the codebase was created to showcase our data's versatility in several applications. We connected some baseline models and other commonly used sets to illustrate better how our data difficulty level compares. In graphs and tables, we denote our dataset as CARLA-BSP.

The code, contained in \texttt{pedestrians-video-2-carla} repository, is written in PyTorch \cite{paszke2019pytorch}, using PyTorch Lightning \cite{falcon2019pytorch} as a framework. We also utilize PyTorch3D \cite{ravi2020accelerating} and CameraTransform \cite{gerum2019cameratransform} libraries to provide crucial functionality.

Quality measures used to evaluate the models are described in Appendix~\ref{apx:measures}.

\subsection{Crossing vs. non-crossing classification}
\label{ssec:examples:classification}

One of the common tasks in analyzing the scene in the automotive domain is deciding if the detected person will cross the street. Depending on the dataset, there are usually two or three classes distinguished. The binary case seems relatively straightforward -- the pedestrian will cross or not. However, even then, everyone needs to agree on what is meant by ,,crossing'' -- walking from one side of the street to the other (even if separate from the future path of the car)? Or is just entering the driving lane enough? Or maybe crossing the planned car trajectory is required? What about the situations in parking lots? What about cyclists? Should they be included, and under what conditions?

When more classes are separated, the usual first criterion seems to be whether the person comes into any interaction with the AV. If not, that pedestrian is labeled as some kind of ,,irrelevant'' class, and only pedestrians with any interaction are separated into the crossing and non-crossing. Both of those approaches to labeling are prone to create class imbalances, as shown in the case of Joint Attention in Autonomous Driving (JAAD) \cite{rasouli2017are,rasouli2017agreeing} and Pedestrian Intention Estimation (PIE) \cite{rasouli2019PIE} datasets. Both datasets can also be used with our codebase, with the ability to specify arbitrary clip lengths and offsets or using the predefined benchmark data modules.

In our dataset, we mark the pedestrian crossing when he enters the driving lane. All other situations are marked as non-crossing. As the scene is recorded from the fixed point, there is no intersection of trajectories to consider. Again, we provide two ways to utilize the data -- the customizable data module (CarlaRecorded) and the benchmark one (CarlaBenchmark), comparable to benchmarks provided for JAAD and PIE \cite{kotseruba2021benchmark}.

The state-of-the-art in pedestrian crossing and non-crossing classification has seen considerable progress, driven by advances in deep learning and diverse datasets. Several approaches have been developed, including temporal models like the SRA-LSTM \cite{yusheng2021sra}, which incorporates social relationship attention mechanisms to improve trajectory prediction. CNN-based methods extract local features from image patches for real-time classification. Spatial attention mechanisms, combine attention with trajectory prediction to enhance understanding of pedestrian behavior. There are also methods based on pose estimation and classification MM-Pose \cite{mmpose2020} which may also fuse multiple data modalities to improve intention estimation. Graph-based methods, such as Social-STGCNN \cite{mohamed2020social}, model interactions between pedestrians and their environment for better insights into crossing behavior. Despite progress, real-world challenges like noisy data, occlusions, and complex interactions persist, emphasizing the need for robust, reliable, and interpretable models in diverse scenarios.

\begin{figure}
    \includegraphics[width=\textwidth]{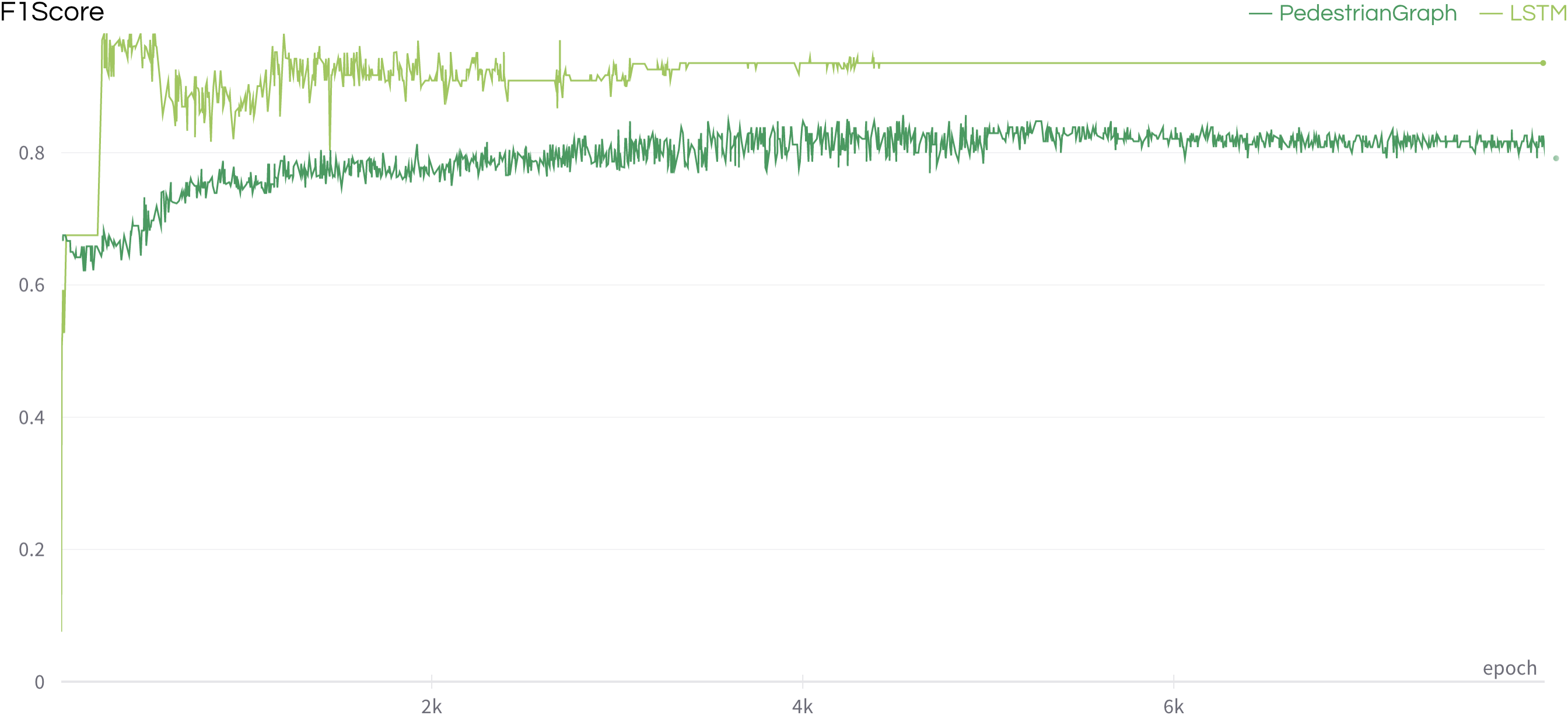}
    \caption{Classification results in terms of F1-score when using perfect CARLA-BSP data. \label{fig:examples:classification:carla-bsp-perfect}}
\end{figure}

We tested the classification flow using the 2D pose keypoints as the input. As an example model, besides the basic LSTM architecture, we adapted the Pedestrian Graph model \cite{cadena2019pedestrian}. We have used the benchmark setup for the experiments, with the clips' length equal to \num{16} frames (a bit over \SI{0.5}{\second}). The overlap between clips was ten frames. We also enabled class balancing during the training, resulting in \num{344} clips in each class. In the first test, the data was perfect, i.e., no artificial noise or missing joints were introduced. To our surprise, in that ideal setup, a simple LSTM model outperformed the PedestrianGraph one (see Fig.~\ref{fig:examples:classification:carla-bsp-perfect}).

When we introduced artificial missing joints and gaussian noise, the performance of the models became closer to that achieved for JAAD and PIE datasets. This performance drop is partially due to the ground truth pose data being incomplete and noisy. According to our tests, in the validation set of the benchmark setup, JAAD is missing \SI{85.74}{\percent} of joints when using data provided in \cite{kotseruba2021benchmark} and \SI{71.87}{\percent} when using data we extracted using OpenPose \cite{cao2021openpose}. For PIE, \SI{93.28}{\percent} keypoints is missing when using data data from \cite{kotseruba2021benchmark} and \SI{84.16}{\percent} in data extracted using OpenPose.

\begin{figure}
    \includegraphics[width=\textwidth]{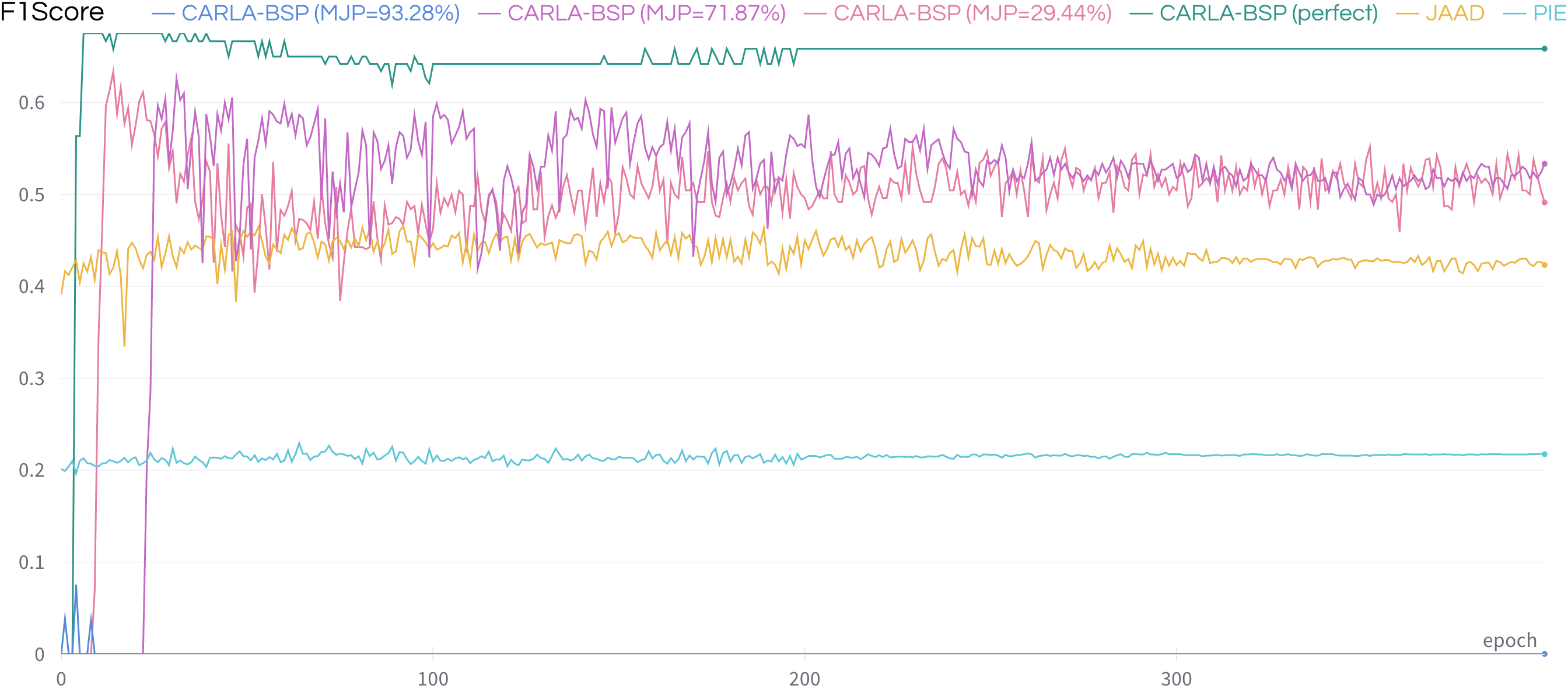}
    \caption{Classification results for short experiments with PedestrianGraph model. \label{fig:examples:classification:pedestrian-graph-f1score}}
\end{figure}

In the short experiments we have run (\num{400} epochs), the performance of the PedestrianGraph model on CARLA-BSP with requested missing joint probability (MJP) \SI{71.87}{\percent} and gaussian noise with STD=5.0 (added to ground truth pixel values), there was an F1-score drop of \num{\approx 0.125}, which placed the results mid-way between the performance for JAAD dataset and perfect CARLA data. Interestingly, a much smaller number of missing joints (\num{\approx 0.3}) resulted in a similar performance. However, when it was increased to \SI{93.28}{\percent}, the PedestrianGraph model failed completely (see Fig.~\ref{fig:examples:classification:pedestrian-graph-f1score}).

\begin{figure}
    \includegraphics[width=\textwidth]{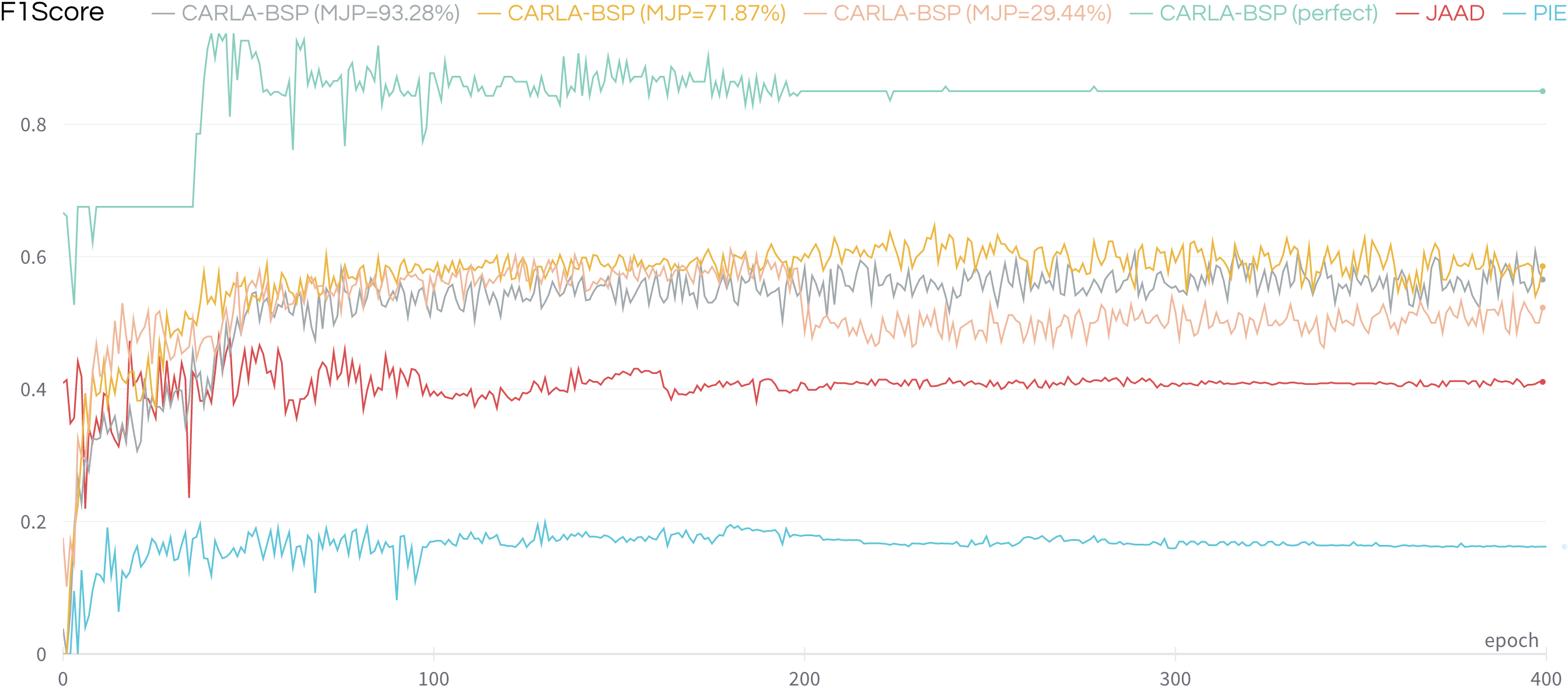}
    \caption{Classification results for short experiments with LSTM model. \label{fig:examples:classification:lstm-f1score}}
\end{figure}

LSTM seems to be more resilient to introducing artificial missing joints -- apart from the initial F1-score drop, the MJP value seems to make a minimal difference. However, it performs slightly worse for JAAD and PIE data (see Fig.~\ref{fig:examples:classification:lstm-f1score}).

\subsection{Pose autoencoder}
\label{ssec:examples:autoencoder}

Autoencoding is a powerful unsupervised learning technique that has emerged as a popular first step in data processing pipelines, owing to its ability to improve data quality and facilitate more effective downstream processing. In the context of the JAAD dataset, autoencoding can be employed to enhance the quality of skeletons extracted using OpenPose, a renowned human pose estimation framework. By optimizing the representation of these skeletons, the aim is to bolster the performance of classification tasks on the JAAD dataset, which consists of annotated video sequences that focus on pedestrian behavior and interactions in urban settings. By leveraging state-of-the-art autoencoding techniques, it is possible to learn a compact and noise-resistant representation of the input data, which can subsequently be used to refine the extracted skeletal features \cite{Carissimi2018FillingTG}. In turn, this refined representation is expected to yield improved classification results, thus demonstrating the value of autoencoding as a preliminary step in processing complex datasets such as JAAD.

Aside from the CARLA-BSP, as the input data we also used CMU \cite{cmuWEB} and HumanEva \cite{sigal2010humaneva} input data, available as a part of AMASS \cite{mahmood2019amass}, as well as JAAD and PIE. We used a simple transformer-based model (with six layers and two heads) that managed to learn clean CARLA-BSP and AMASS data almost perfectly. 

\begin{figure}
    \includegraphics[width=\textwidth]{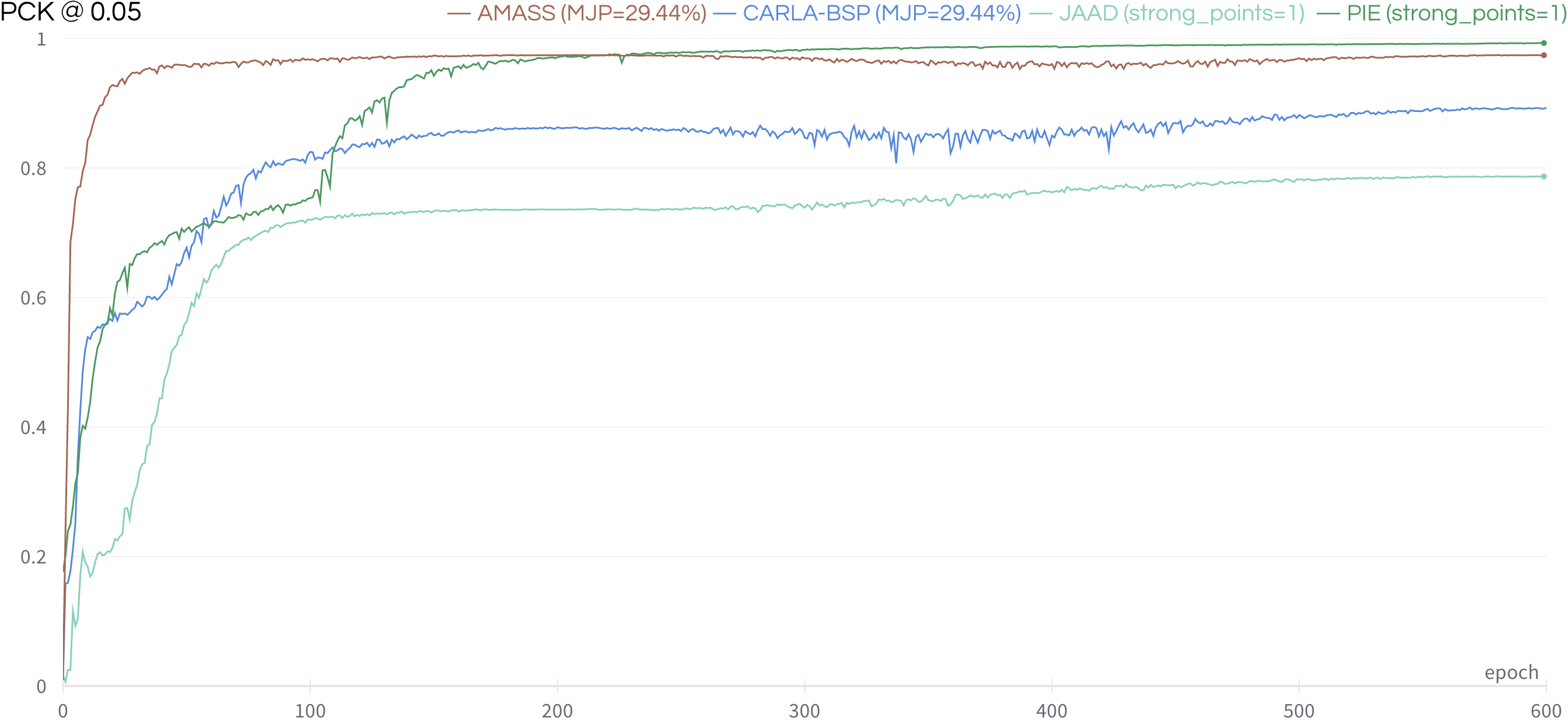}
    \caption{PCK@0.05 metrics during training for simple transformer-based autoencoder and noisy CARLA-BSP and AMASS data contrasted with JAAD@1.0 and PIE@1.0 runs. \label{fig:examples:autoencoder:noisy-pck}}
\end{figure}

\begin{figure}
    \includegraphics[width=\textwidth]{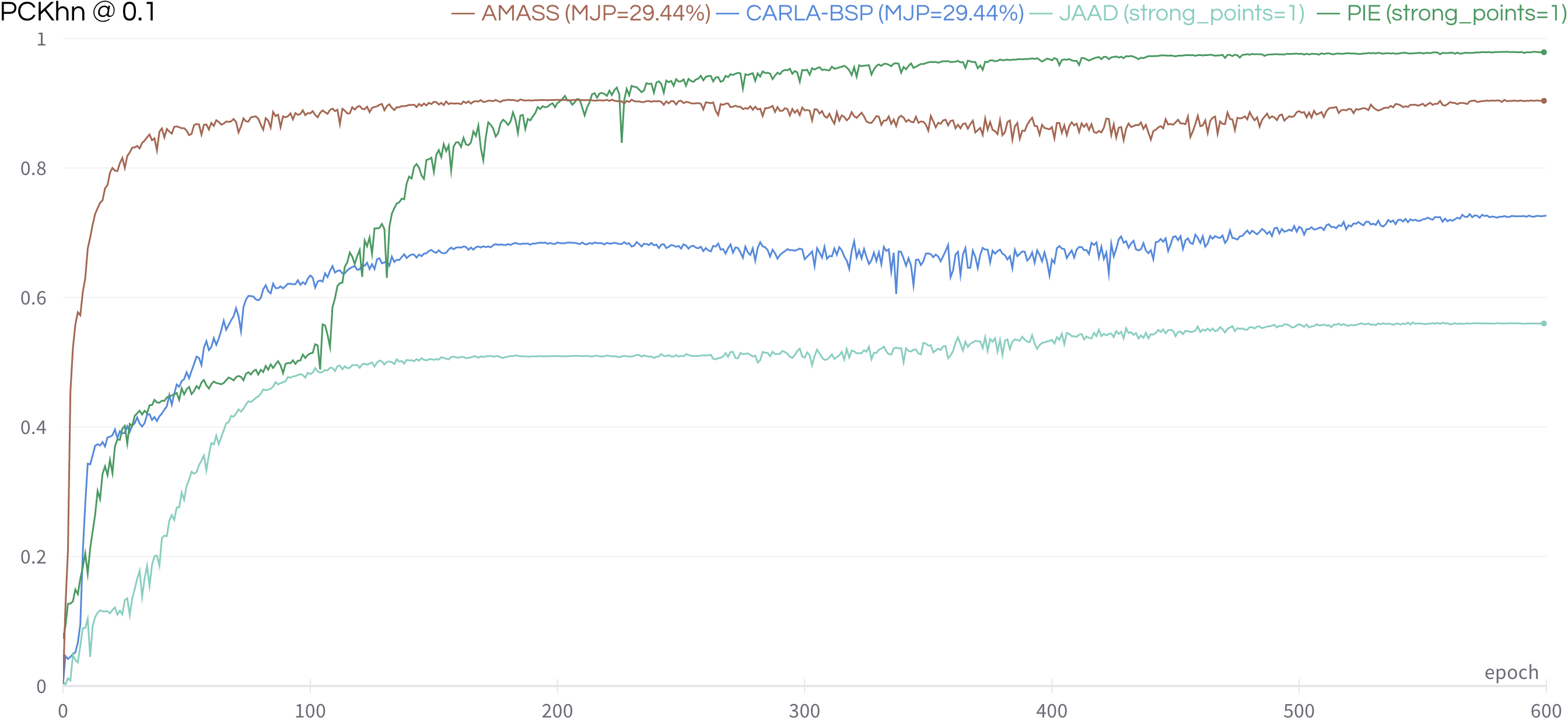}
    \caption{PCKhn@0.1 metrics during training for simple transformer-based autoencoder and noisy CARLA-BSP and AMASS data contrasted with JAAD@1.0 and PIE@1.0 runs. \label{fig:examples:autoencoder:noisy-pckhn}}
\end{figure}

The first conclusion we reached is a confirmation, that the amount of available data plays a crucial role in the learning process. It can be observed that while for the perfect data (CARLA-BSP and AMASS), the convergence time and end quality of the model are very similar, the introduction of the same amount of gaussian noise (STD=5.0) and artificial missing joints (MJP=\SI{29.44}{\percent}) drastically changes the results (Fig.~\ref{fig:examples:autoencoder:noisy-pck} and \ref{fig:examples:autoencoder:noisy-pckhn}). The run using AMASS dataset, containing \num{\approx 10} times more training data, approached the ideal results. In contrast, the one for CARLA-BSP resulted in a performance drop of \num{\approx 10} percentage points (p.p.) in terms of PCK@0.05 (percentage of correct keypoints with a tolerance of \SI{5}{\percent} of bounding box; values are between 0.0 and 1.0) and almost \num{18} p.p. drop in terms of PCKhn@0.1 (percentage of correct keypoints with a tolerance of \SI{10}{\percent} of hips-neck distance; values are between 0.0 and 1.0).

This effect can also be noticed when comparing the runs on a strong\_points=1.0 subset of PIE and JAAD datasets, where 'strong points' denote the desired percentage of non-zero, hopefully semi-correctly detected, keypoints in each frame (denoted as PIE@1.0 and JAAD@1.0, respectively). Run for PIE@1.0 eventually approaches the perfect metrics, while the one for JAAD@1.0 reaches plateau at PCK@0.05 \num{\approx 0.79} and PCKhn@0.1 \num{\approx 0.56}. The subset of PIE@1.0 data is \num{\approx 3} times bigger than the one for JAAD@1.0.

The second conclusion is that imperfections introduced during the pose estimation process seem easier for the model to correct than the artificial noise. This effect can be observed when comparing the PIE@1.0 run (which eventually approaches perfect results) with the CARLA-BSP mentioned before (Fig.~\ref{fig:examples:autoencoder:noisy-pck}), which resulted in a severe performance drop even though CARLA-BSP has over \num{14} times more training data than the PIE@1.0 subset!

\begin{figure}
    \includegraphics[width=\textwidth]{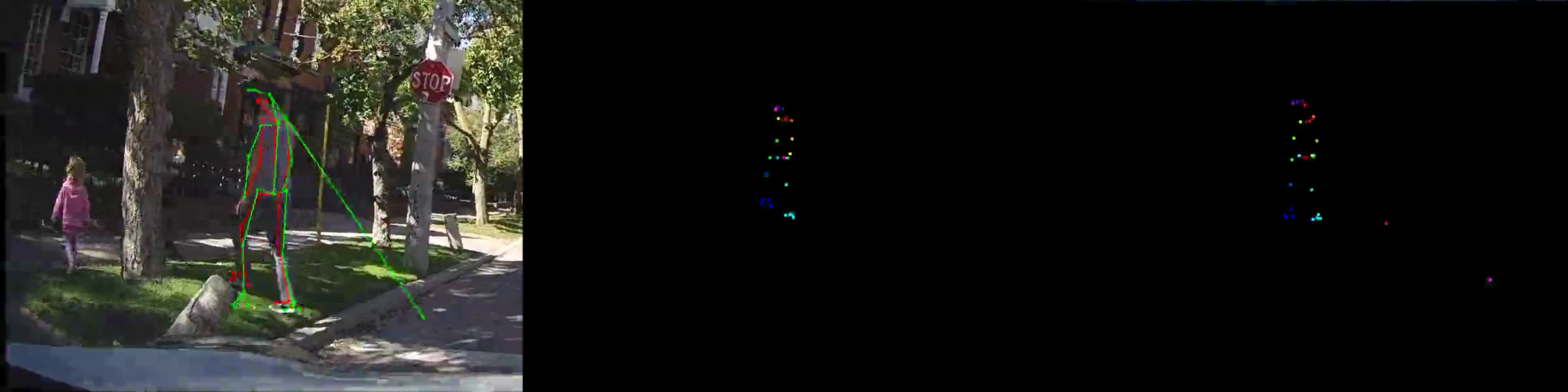}
    \includegraphics[width=\textwidth]{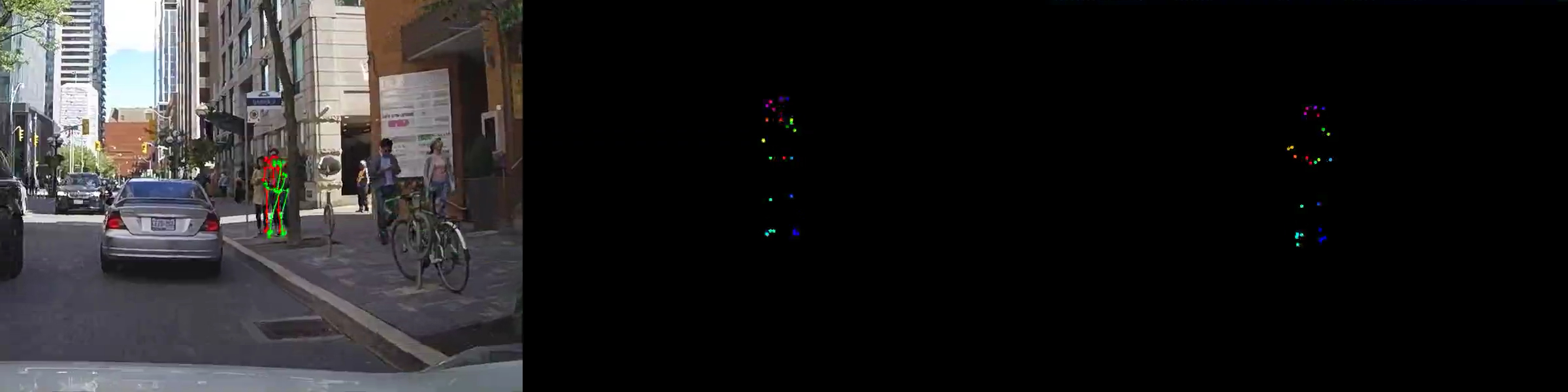}
    \caption{Example PIE@0.1 rendering results. From left: fragment of a source video with original pose data (red) and model output (green), normalized input pose data as dots, model output as dots. \label{fig:examples:autoencoder:pie-render}}
\end{figure}

\begin{figure}
    \includegraphics[width=\textwidth]{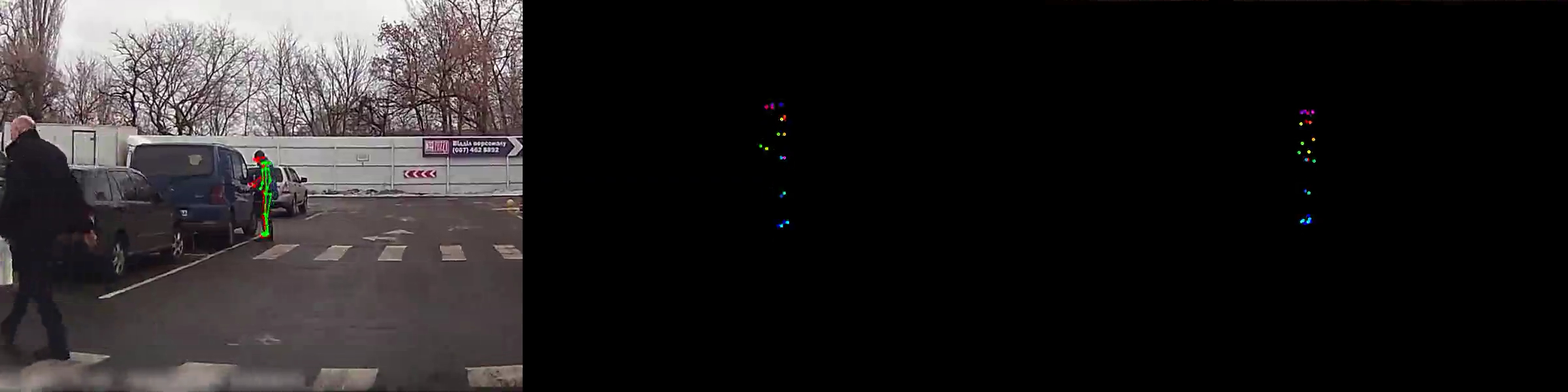}
    \includegraphics[width=\textwidth]{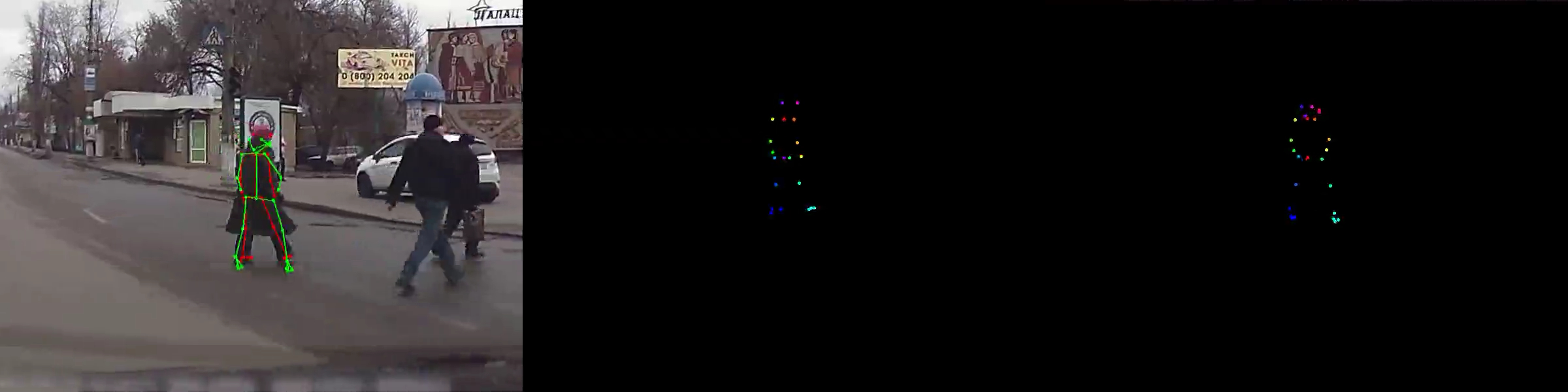}
    \caption{Example JAAD@0.1 rendering results. From left: fragment of a source video with original pose data (red) and model output (green), normalized input pose data as dots, model output as dots. \label{fig:examples:autoencoder:jaad-render}}
\end{figure}

The third conclusion is that training our simple transformer-based autoencoder only on very incomplete PIE@0.1 or JAAD@0.1 data (at least \SI{10}{\percent} of data visible in each frame) yields no usable results (see Fig.~\ref{fig:examples:autoencoder:pie-render} and Fig.~\ref{fig:examples:autoencoder:jaad-render} for example outputs). The maximal values of PCK@0.05 are \num{\approx0.5}, and PCKhn@0.1 \num{\approx0.3}. In both cases, the model learns some approximation of the pose. However, it significantly shifts some keypoints. Sometimes, the returned pose seems much closer to some averaged pose and does not follow the actual person's movements, especially in the JAAD@0.1 case.

\begin{figure}
    \parbox[c]{0.8\textwidth}{
        \raisebox{-0.5\height}{1}\hspace*{12pt}\raisebox{-0.5\height}{\includegraphics[width=0.74\textwidth]{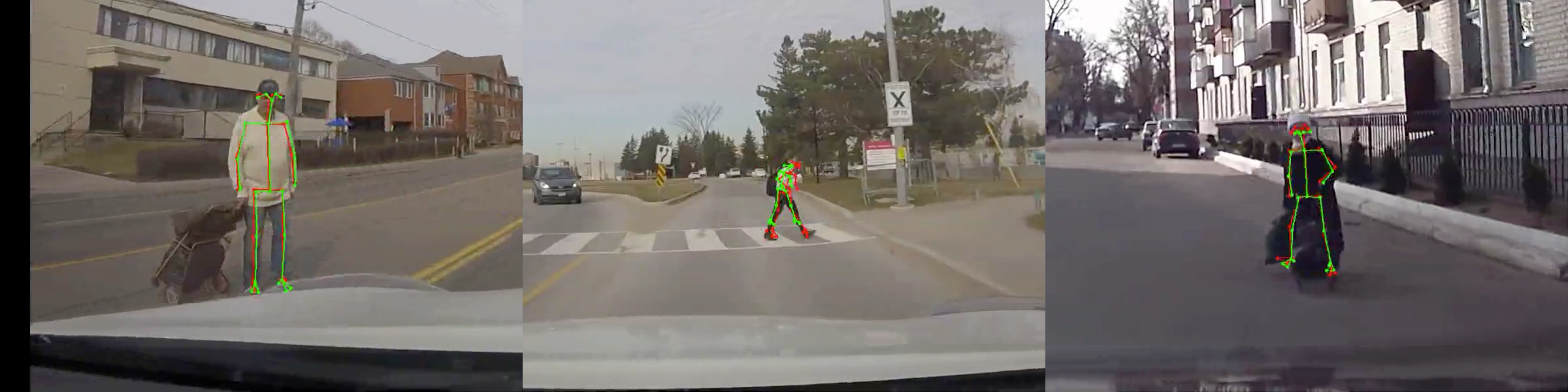}}\\
        
        \raisebox{-0.5\height}{2}\hspace*{12pt}\raisebox{-0.5\height}{\includegraphics[width=0.74\textwidth]{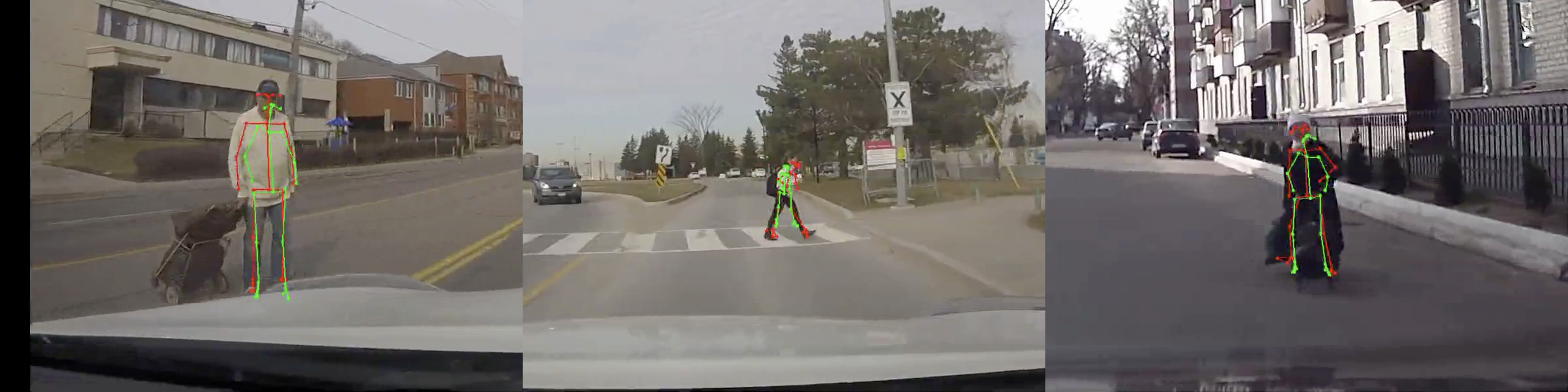}}\\
        
        \raisebox{-0.5\height}{3}\hspace*{12pt}\raisebox{-0.5\height}{\includegraphics[width=0.74\textwidth]{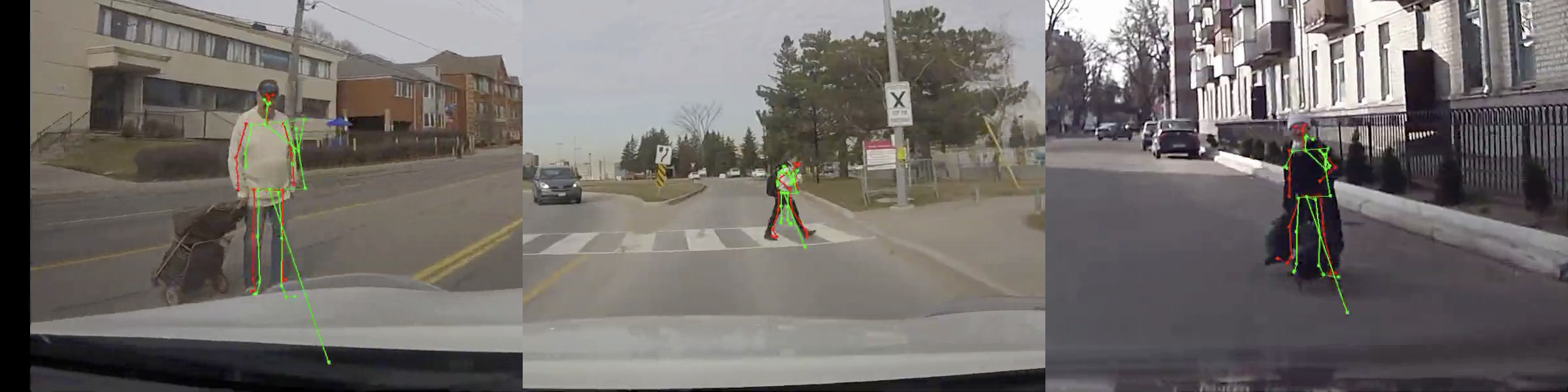}}\\
        
        \raisebox{-0.5\height}{4}\hspace*{12pt}\raisebox{-0.5\height}{\includegraphics[width=0.74\textwidth]{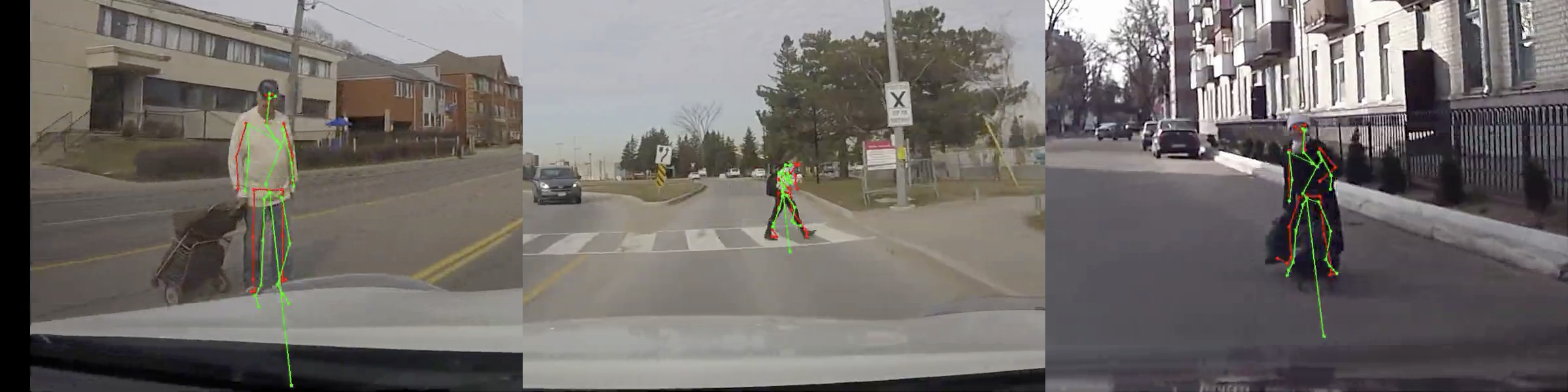}}\\
        
        \raisebox{-0.5\height}{5}\hspace*{12pt}\raisebox{-0.5\height}{\includegraphics[width=0.74\textwidth]{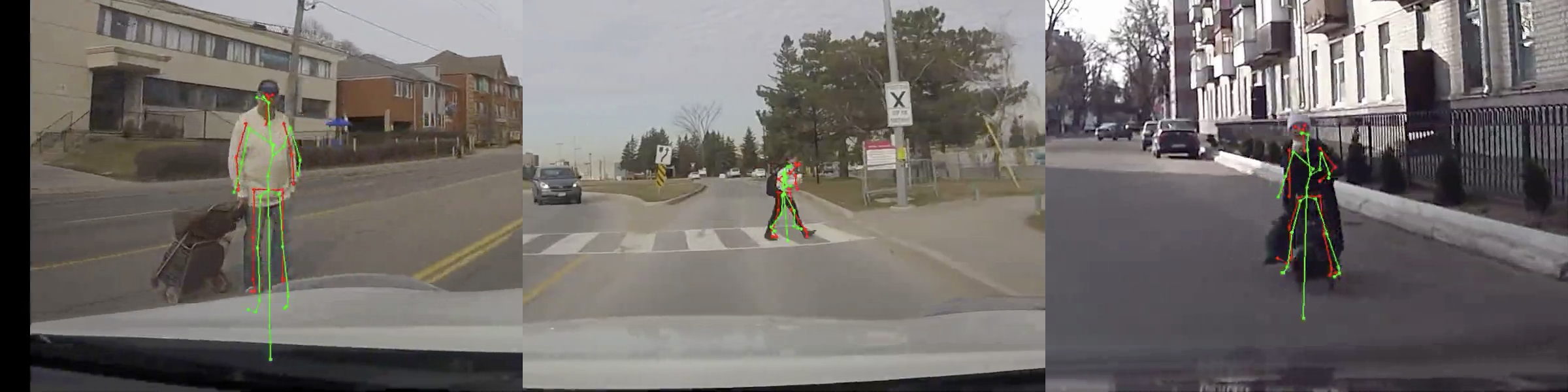}}\\
        
        \raisebox{-0.5\height}{6}\hspace*{12pt}\raisebox{-0.5\height}{\includegraphics[width=0.74\textwidth]{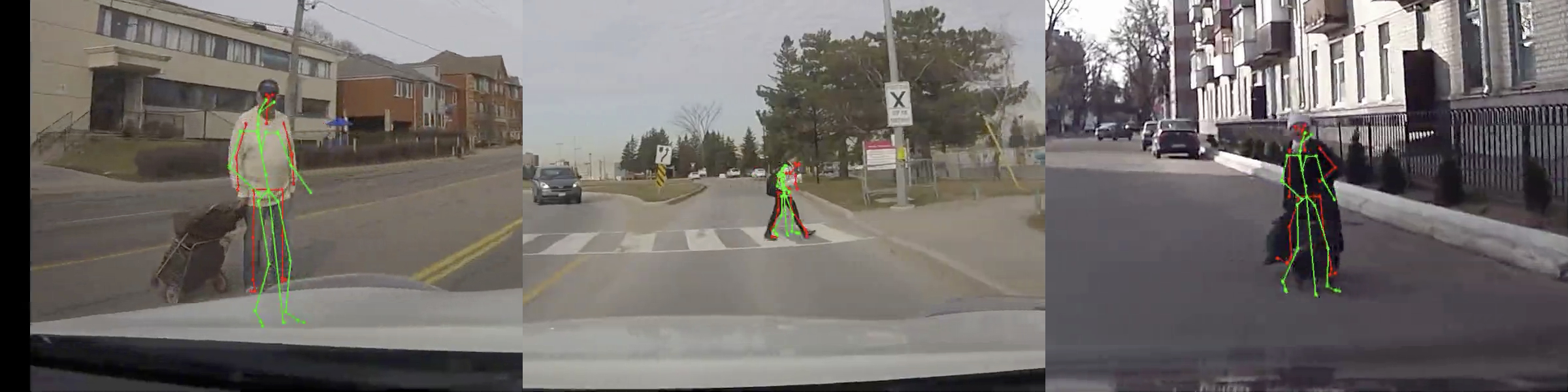}}\\
        
        \raisebox{-0.5\height}{7}\hspace*{12pt}\raisebox{-0.5\height}{\includegraphics[width=0.74\textwidth]{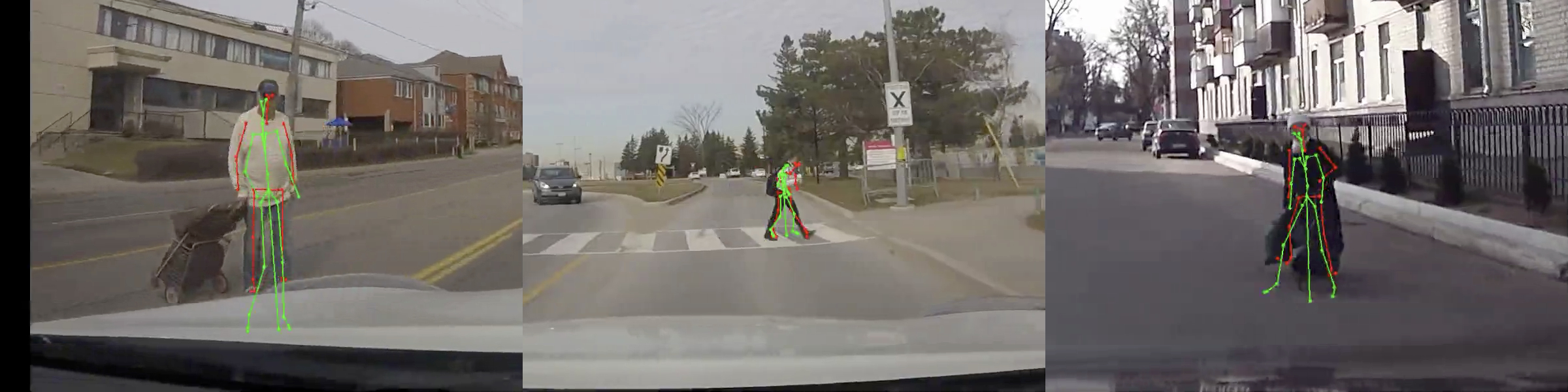}}
    }%
    \parbox{0.2\textwidth}{
        \raggedleft
        \hypersetup{urlcolor=black}
        \qrcode[height=60pt]{https://youtu.be/SZzR7CNjtlQ?utm_source=QRCode&utm_medium=latex&utm_campaign=CARLA-BSP}\\
        
        \vspace{36pt}
        
        \qrcode[height=60pt]{https://youtu.be/k9zPpkoWJhk?utm_source=QRCode&utm_medium=latex&utm_campaign=CARLA-BSP}\\
        
        \vspace{36pt}
        
        \qrcode[height=60pt]{https://youtu.be/6GJB-uLEpoo?utm_source=QRCode&utm_medium=latex&utm_campaign=CARLA-BSP}
    }
    \caption{Example rendering results for JAAD@1.0 test set. Order (from the top) follows rows in Tab.~\ref{tab:examples:autoencoder:mixed}. Follow QR codes to see videos for each pedestrian. \label{fig:examples:autoencoder:mixed-render}}
\end{figure}

In the next step, we trained the model on mixed datasets, meaning that the model was shown data from either JAAD, CARLA-BSP or AMASS in a single epoch. To make it possible, we mapped the various source pose data to common format (CARLA skeleton one), which had a side effect of increasing the number of missing joints and noise. This effect occurred because we only copied the values from the existing close joints, without any additional processing. Then, we tested all models on the same strong\_points=1.0 subset of JAAD (JAAD@1.0) and strong\_points=1.0 subset of PIE (PIE@1.0). The results can be seen in Tab.~\ref{tab:examples:autoencoder:mixed} and example rendered frames in the Fig.~\ref{fig:examples:autoencoder:mixed-render}. It is important to note however, that since JAAD@1.0 and JAAD@0.1 dataset variants were generated independently, models trained using JAAD@0.1 have seen about $\nicefrac{1}{3}$ of JAAD@1.0 test data during training, which likely inflates the results.

\begin{table}
    \scriptsize
    \setlength{\tabcolsep}{2.5pt}
    \setlength{\cmidrulekern}{2.5pt}
    \begin{tabularx}{\textwidth}{c>{\raggedright}X*{7}c}
         \toprule
         & \multicolumn{4}{c}{\textbf{Train}} & \multicolumn{2}{c}{\textbf{Test JAAD@1.0}} & \multicolumn{2}{c}{\textbf{Test PIE@1.0}} \\
         \cmidrule(r){2-5}\cmidrule(lr){6-7}\cmidrule(l){8-9}
         & \textbf{Dataset} & \textbf{Proportions} & \textbf{Set size} & \textbf{Skeleton} & \textbf{PCK@0.05} & \textbf{PCKhn@0.1} & \textbf{PCK@0.05} & \textbf{PCKhn@0.1} \\
         \midrule
         1 & JAAD@1.0 & \SI{100}{\percent} & \num{220} & BODY\_25 & \num{0.7635} & \num{0.5581} & \num{0.7865} & \num{0.5558} \\
         2 & JAAD@0.1 & \SI{100}{\percent} & \num{8822} & BODY\_25 & \num{0.4825}* & \num{0.2642}* & \num{0.4965} & \num{0.2605} \\
         3 & JAAD@1.0 / CARLA-BSP / AMASS; MJP=\SI{29.44}{\percent} + gaussian noise (STD=5.0) for all datasets & \SI{10}{\percent}/\SI{40}{\percent}/\SI{50}{\percent} & \num{2200} & CARLA & \num{0.4898} & \num{0.3214} & \num{0.4246} & \num{0.2414} \\
         4 & JAAD@0.1 / CARLA-BSP / AMASS; no~noise & \SI{10}{\percent}/\SI{40}{\percent}/\SI{50}{\percent} & \num{21556} & CARLA & \num{0.4973}* & \num{0.3214}* & \num{0.5352} & \num{0.3019} \\
         5 & JAAD@0.1 / CARLA-BSP / AMASS; MJP=\SI{29.44}{\percent} + gaussian noise (STD=5.0) for CARLA-BSP and AMASS datasets & \SI{10}{\percent}/\SI{40}{\percent}/\SI{50}{\percent} & \num{21556} & CARLA & \num{0.3859}* & \num{0.2240}* & \num{0.5048} & \num{0.3152} \\
         6 & CARLA-BSP; MJP=\SI{29.44}{\percent} + gaussian noise (STD=5.0) & \SI{100}{\percent} & \num{8623} & CARLA & \num{0.2146} & \num{0.1130} & \num{0.2162} & \num{0.1063} \\
         7 & CARLA-BSP / AMASS; MJP=\SI{29.44}{\percent} + gaussian noise (STD=5.0) for all datasets & \SI{50}{\percent} / \SI{50}{\percent} & \num{17246} & CARLA & \num{0.1806} & \num{0.1141} & \num{0.1749} & \num{0.1094} \\
         \bottomrule
    \end{tabularx}

    \vspace{2pt}
    \noindent
    {\footnotesize* ca. $\nicefrac{1}{3}$ of the test data was seen during training.}
    
    \caption{Setup and results for the tests of models trained using various datasets. \label{tab:examples:autoencoder:mixed}}
\end{table}

It can be seen that training on JAAD@1.0 yields the best results when testing on JAAD@1.0 and PIE@1.0. This may be due to the compatibility of the source points data and the fact that there were no missing points in training. However, when trained with missing data (JAAD@0.1) the results were significantly worse. Interestingly, the visual examination of the results (see Fig.~\ref{fig:examples:autoencoder:mixed-render} and linked videos) suggests that while autoencoder trained only on CARLA-BSP / AMASS dataset combination (without seeing any JAAD data) scored near the bottom in terms of PCK@0.05 and PCKhn@0.1, the generated pose is often an approximation of the correct one. It seems to be deformed mainly due to differences in points position when converted between BODY\_25 and CARLA skeletons -- the difference between hips and neck positions makes the denormalized pose taller than the actual one. On the other hand, using an autoencoder trained only on CARLA-BSP seems to result in poses that are mostly turned right from the observer's point of view. This suggests an existence of a bias in the training data, which should be eliminated in future CARLA-based dataset versions.

\subsection{Pose estimation}
\label{ssec:examples:pose-estimation}

Pose estimation is another problem that can benefit from simulated data. Historically, monocular images and videos were used for 2D pose estimation, and 3D required specialized datasets obtained via motion capture technology or at least stereoscopic images. Nowadays, many models offer at least basic 3D pose estimation from monocular images and videos. Example pose estimation solutions include OpenPose \cite{cao2021openpose} mentioned before, UniPose \cite{artacho2020unipose}, AlphaPose \cite{li2021hybrik}, SMPLify-X \cite{pavlakos2019expressive} and others.

The advancements in 3D pose estimation from monocular images and videos have been facilitated by the development of deep learning techniques, specifically convolutional neural networks (CNNs) and their variants. These models have been successful in capturing complex spatial dependencies and learning rich feature representations from 2D images, which have proven beneficial for the task of pose estimation. Additionally, the availability of large-scale annotated datasets, such as MPII \cite{andriluka14cvpr}, and Human3.6M \cite{h36m_pami}, has provided researchers with the necessary data to train and fine-tune these models, resulting in improved performance on both 2D and 3D pose estimation tasks.

Despite the progress made, pose estimation still faces several challenges, such as occlusions, varying illumination, and diverse body shapes and clothing. Simulated data can help address these challenges by providing a controlled environment in which researchers can generate large amounts of annotated data, manipulate object properties, and create a variety of scenarios that are difficult or expensive to obtain with real-world data. For example, synthetic human models can be generated with different body shapes, clothing, and poses, while the virtual environment can be altered to simulate different lighting conditions and occlusion patterns. This can lead to more robust and generalizable models, as they are exposed to a broader range of scenarios during training. Furthermore, simulated data can help overcome the limitations of motion capture technology and stereoscopic images, which are often costly and time-consuming to acquire. By leveraging simulated data, researchers can accelerate the development of pose estimation solutions and contribute to the advancement of the field.

\begin{figure}[tbp]
    \includegraphics[width=\textwidth]{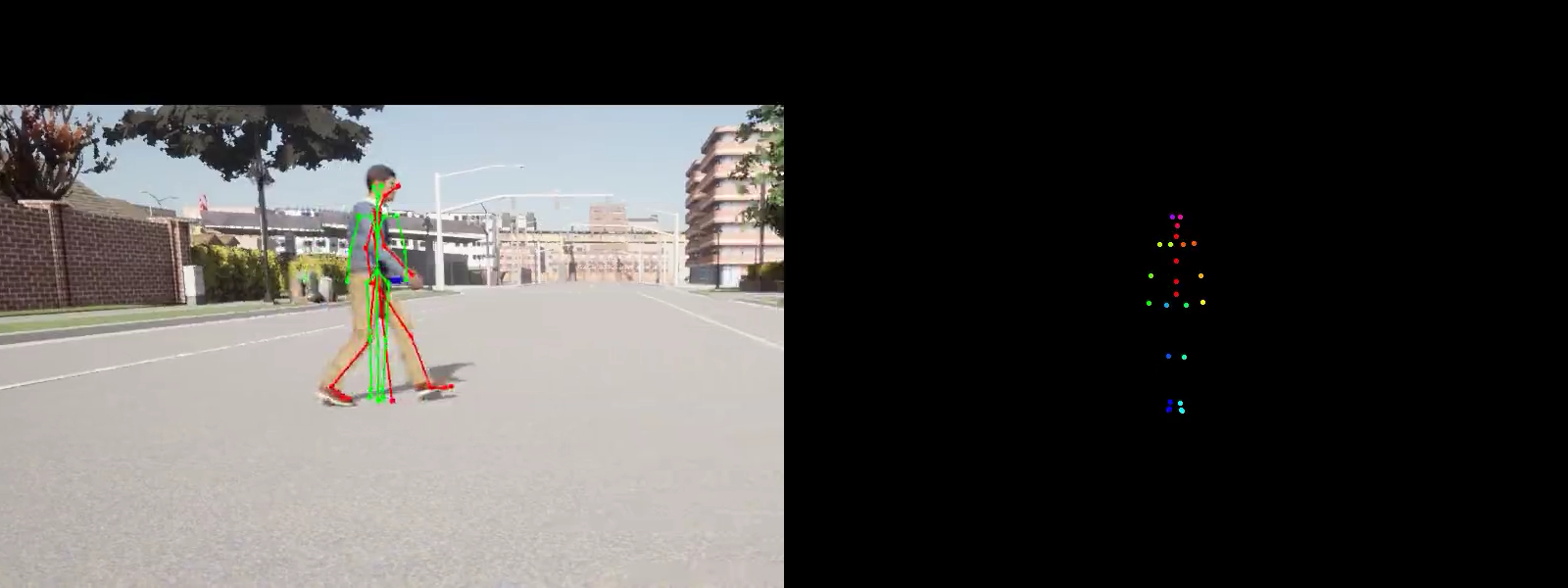}
    \includegraphics[width=\textwidth]{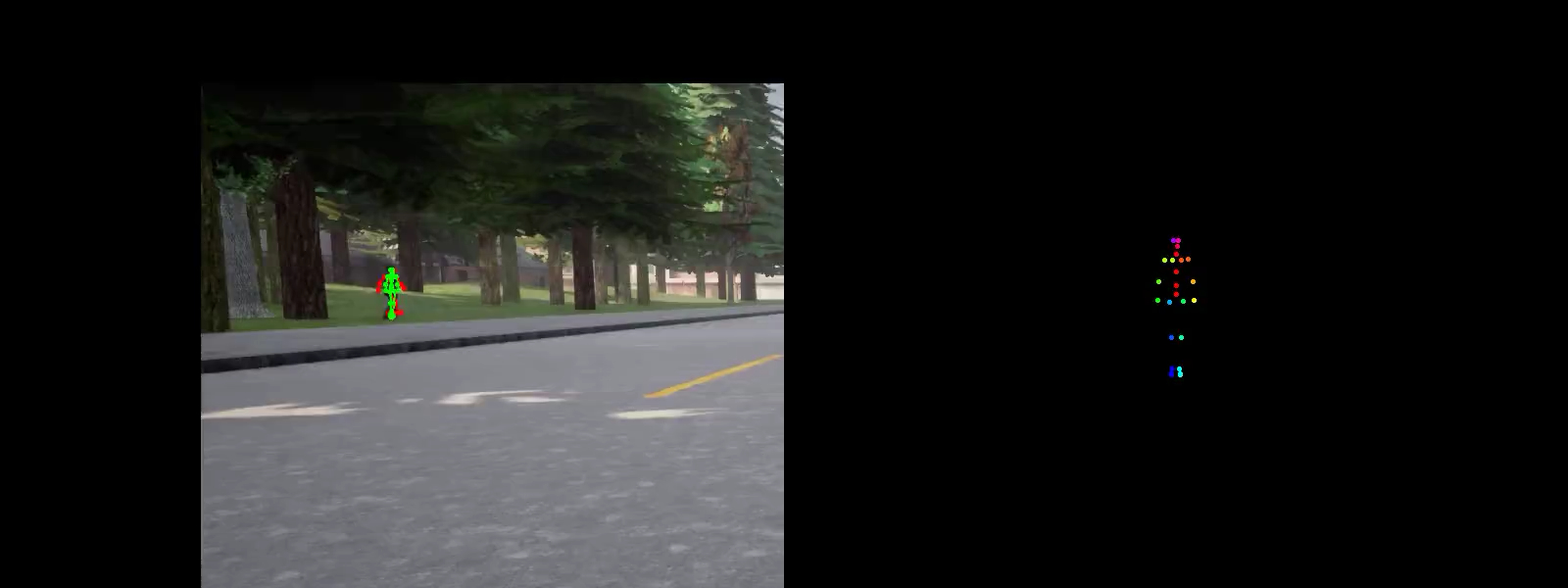}
    \caption{Visualization of pose estimation results. It can be seen that the model learned the average skeleton. \label{fig:examples:pose-estimation:average-skeleton}}
\end{figure}

Due to the limitations of computing power, we could not achieve any meaningful results in the 2D pose estimation task. The models we ran reached a plateau of about \num{0.5} PCK@0.05 and \num{0.35} PCKhn@0.1. The metrics are calculated based on normalized skeletons aligned by hips, so they do not verify if the model even managed to find a pedestrian in the frame. In practice, those results mean that model learned the average skeleton position, as seen in Fig.~\ref{fig:examples:pose-estimation:average-skeleton}.

\subsection{Pose lifting}
\label{ssec:examples:pose-lifting}

Pose lifting can be considered a subset of the pose estimation tasks. In some solutions, when trying to obtain a 3D pose, the 2D pose is estimated first, and then, based on the 3D pose, it is lifted to 3D. Some of the example solutions include 3DPoseBaseline \cite{martinez2017simple} and PoseFormer \cite{zheng20213dhuman}. We trained both of those models, as well as LSTM, using the CARLA-BSP dataset. All models were trained using the standard 2D (x,y) $\rightarrow$ 3D (x,y,z) mapping that we refer to as the 'absolute\_loc` model output type. For the LSTM, we also tested the 'relative\_rot' mode -- given the 2D (x,y) points, we expected the model to learn the rotations of each joint relative to its parent in the kinematic tree.

\begin{figure}
    \includegraphics[width=\textwidth]{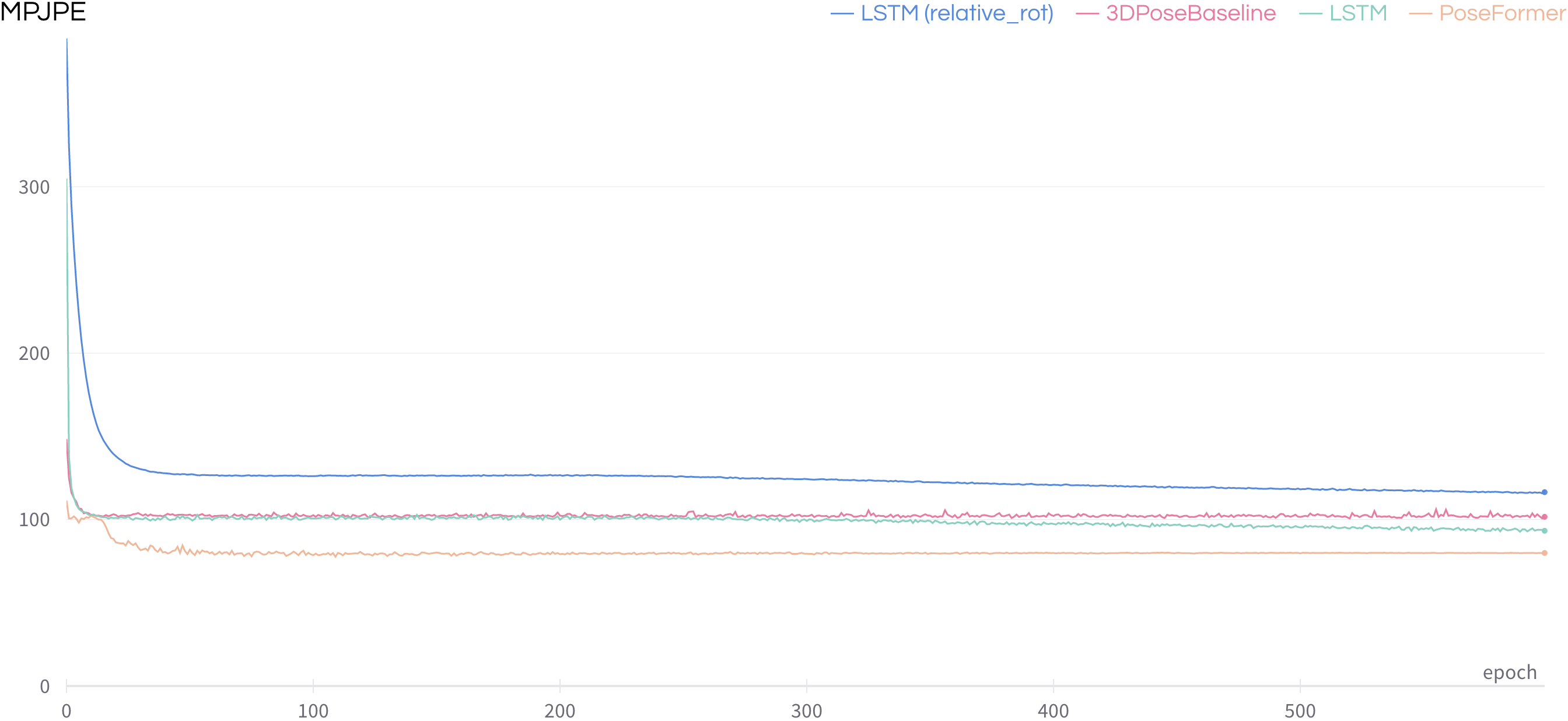}
    \caption{Change in MPJPE calculated on validation subset during the training. \label{fig:examples:pose-lifting:training}}
\end{figure}

\begin{figure}
    \centering
    \includegraphics[width=\textwidth]{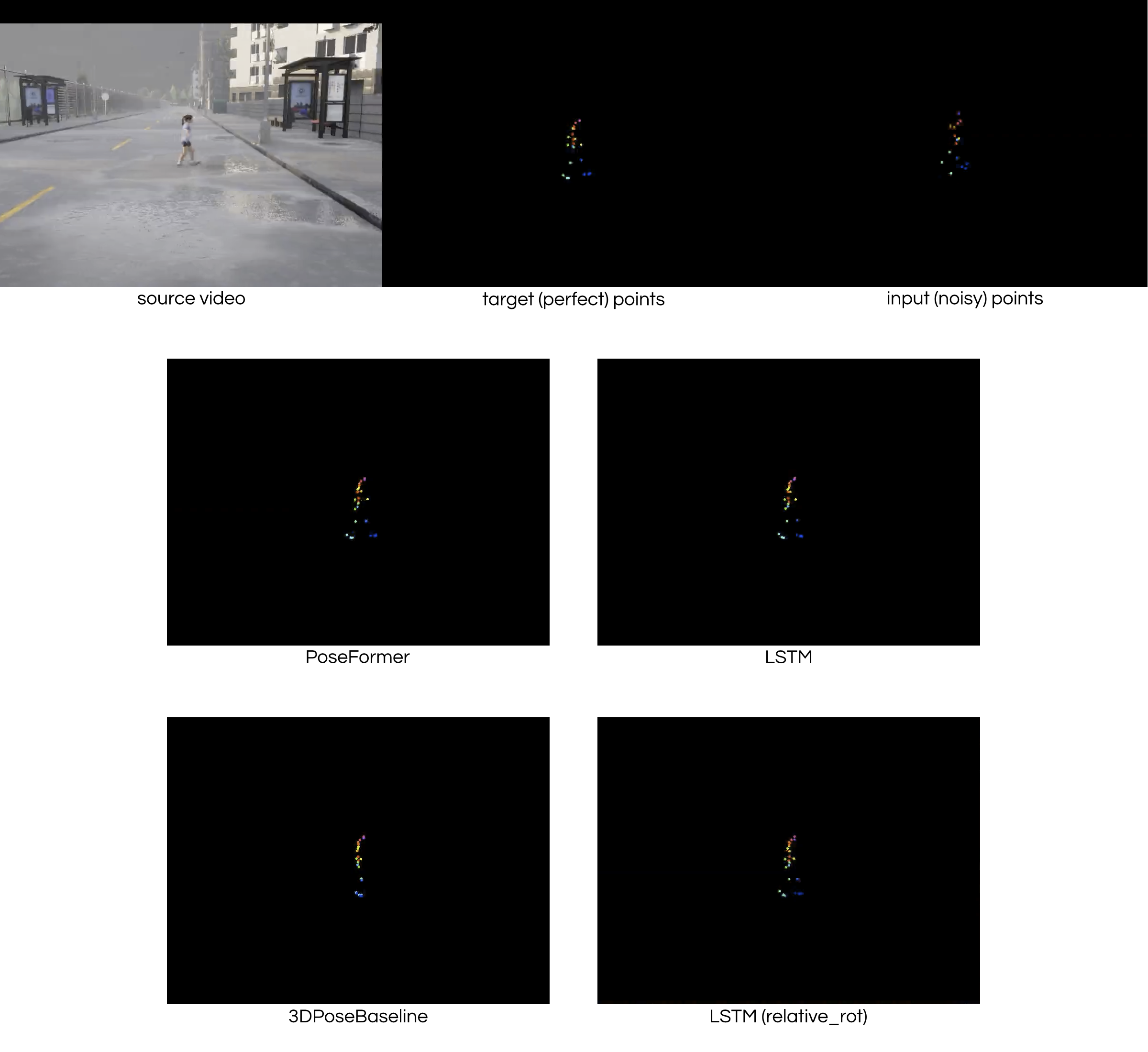}
    {
        
        \parbox{60pt}{%
            \hypersetup{urlcolor=black}%
            \centering%
            \qrcode[height=60pt]{https://youtu.be/Pv-2bmwHNTc?utm_source=QRCode&utm_medium=latex&utm_campaign=CARLA-BSP}\\%
            \vspace{6pt}%
            \scriptsize\fontfamily{qag}\selectfont\linespread{1} PoseFormer%
        }\hspace{36pt}%
        \parbox{60pt}{%
            \hypersetup{urlcolor=black}%
            \centering%
            \qrcode[height=60pt]{https://youtu.be/PM7B2ekQ5mo?utm_source=QRCode&utm_medium=latex&utm_campaign=CARLA-BSP}\\%
            \vspace{6pt}%
            \scriptsize\fontfamily{qag}\selectfont\linespread{1} LSTM%
        }\hspace{36pt}%
        \parbox{60pt}{%
            \hypersetup{urlcolor=black}%
            \centering%
            \qrcode[height=60pt]{https://youtu.be/POus_q63w-Q?utm_source=QRCode&utm_medium=latex&utm_campaign=CARLA-BSP}\\%
            \vspace{6pt}%
            \scriptsize\fontfamily{qag}\selectfont\linespread{1} 3DPoseBaseline%
        }\hspace{36pt}%
        \parbox{60pt}{%
            \hypersetup{urlcolor=black}%
            \centering%
            \qrcode[height=60pt]{https://youtu.be/u8g8E30Yt3E?utm_source=QRCode&utm_medium=latex&utm_campaign=CARLA-BSP}\\%
            \vspace{6pt}%
            \scriptsize\fontfamily{qag}\selectfont\linespread{1} LSTM (relative\_rot)%
        }
    }

    \caption{Example outputs from pose lifting models. In visualization videos, all model outputs are vertically flipped. Follow QR codes to see more examples. \label{fig:examples:pose-lifting:outputs}}
\end{figure}

We capped the training at \num{600} epochs. The models reached close-to-final performance after ca. \num{100} epochs, although LSTM-based ones were still showing small improvements until the end of training (see Fig.~\ref{fig:examples:pose-lifting:training}). As the input, all models get 2D points normalized with hips-neck distance. The final results, calculated on the test set, can be seen in Tab.~\ref{tab:examples:pose-lifting:test} and the example outputs in Fig.~\ref{fig:examples:pose-lifting:outputs}. We also experimented with loss functions; however, out of all that we tested, the classic MSE calculated using normalized 3D input points and model output was the most effective.

\begin{table}
    \scriptsize
    \begin{tabularx}{\textwidth}{*{6}X}
         \toprule
         Model & Output type & MPJPE [\si{\milli\meter}] & N-MPJPE [\si{\milli\meter}] & PA-MPJPE [\si{\milli\meter}] & MPJVE [\si{\milli\meter}] \\
         \midrule
         PoseFormer     & absolute\_loc & \num{85.54}   & \num{82.855}  & \num{23.563} & \num{26.218} \\
         LSTM           & absolute\_loc & \num{109.359} & \num{107.939} & \num{64.029} & \num{29.567} \\
         3DPoseBaseline & absolute\_loc & \num{119.067} & \num{118.388} & \num{76.965} & \num{29.043} \\
         LSTM           & relative\_rot & \num{131.905} & \num{131.871} & \num{99.412} & \num{29.596} \\
         \bottomrule
    \end{tabularx}

    \caption{Quality measures in pose lifting task calculated using a test set of CARLA-BSP. \label{tab:examples:pose-lifting:test}}
\end{table}

\subsection{Other possible applications}
\label{ssec:examples:other}

Apart from the described cases, the synthetic datasets generated from CARLA can also be used in other tasks. Some of them are obvious and well-researched, like pedestrian (or people in general) detection and tracking. Others may include scene semantic segmentation, motion retargeting, or style transfer. We hope that increasing the complexity and realism of the generated pedestrians and the world around them can open doors to even more exciting possibilities, for example:

\begin{itemize}
\item \textbf{Advanced pedestrian behavior modeling:} As AI and deep learning techniques evolve, CARLA-based datasets can help create sophisticated pedestrian behavior models, predicting their actions in real-time, resulting in safer driving decisions by autonomous vehicles.

\item \textbf{Adversarial training:} The setup can be leveraged to create adversarial scenarios that challenge the robustness of autonomous vehicle systems. By exposing AI models to a variety of difficult situations, developers can ensure that self-driving cars respond appropriately to extreme or uncommon events.

\item \textbf{Multi-modal sensor fusion:} By incorporating various sensor modalities like RGB cameras, DVS cameras, LIDAR, and others, the setup enables the development of algorithms that fuse data from multiple sources for more accurate perception and decision-making.

\item \textbf{Ethics and AI decision-making:} The setup can be employed to develop and test AI decision-making algorithms that consider ethical dilemmas and incorporate human values, striking a balance between safety, efficiency, and fairness in real-world scenarios.
\end{itemize}

\section{Conclusions and future work}

The generation of synthetic datasets for pedestrian crossing and non-crossing detection problems offers a multitude of advantages, particularly in the context of multimodal sensing and the increasing complexity of real-world scenes. One of the primary benefits of synthetic datasets is their ability to capture a diverse range of scenarios, environmental conditions, and pedestrian behaviors that may not be readily available in real-world datasets. This is crucial for training robust machine learning models capable of handling the variability and intricacies of pedestrian crossing situations. 

A key aspect of our future work will involve integrating multiple sensor modalities, such as LiDAR, cameras, and radar, to form a holistic representation of the environment. By fusing the complementary strengths of these sensing modalities, we aim to build more robust and accurate pedestrian detection models that can seamlessly adapt to the dynamic nature of urban environments. The multimodal approach will also aid in addressing challenges related to occlusions, varying lighting conditions, and other complexities commonly encountered in real-world pedestrian crossing scenarios.

\section*{Funding}

\begin{minipage}{\textwidth} 
\includegraphics[width=\textwidth]{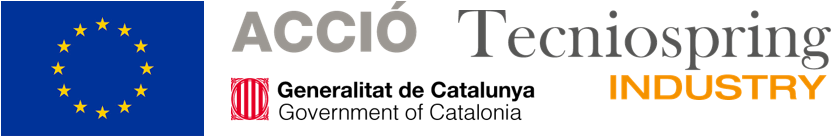}
\noindent
The project on which these results are based has received funding from the European
Union's Horizon 2020 research and innovation programme under Marie Skłodowska-Curie
grant agreement No. 801342 (Tecniospring INDUSTRY) and the Government of Catalonia's
Agency for Business Competitiveness (ACCIÓ).
\end{minipage}
\vspace{0.001pt}

This work only expresses the opinion of the author and neither the European
Union nor ACCIÓ are liable for the use made of the information provided.

\appendix

\section{Quality measures}
\label{apx:measures}

\subsection{Classification}
\label{apx:measures:classification}

In binary classification tasks usually we are interested only in the model performance for the positive class, which in our case means that a pedestrian is going to cross. Due to class imbalance inherent to most of the used datasets, the standard accuracy measure is meaningless. Therefore, we only report the results using positive-class F1-score, which is a compound metric that combines precision and recall (also known as sensitivity).

\begin{equation}
    \mathrm{recall} = \frac{\mathit{tp}}{\mathit{tp} + \mathit{fn}},
    \label{eq:recall}
\end{equation}
\begin{equation}
    \mathrm{precision} = \frac{\mathit{tp}}{\mathit{tp} + \mathit{fp}},
    \label{eq:precision}
\end{equation}
where:
\begin{itemize}
    \item $\mathit{tp}$ -- true positive -- item correctly classified as belonging to positive class,
    \item $\mathit{fp}$ -- false positive -- item incorrectly classified as belonging to positive class,
    \item $\mathit{fn}$ -- false negative -- item incorrectly classified as belonging to negative class.
\end{itemize} 

The $\beta$ parameter controls the recall importance in relevance to the precision when calculating an $\mathrm{F}_\beta$-measure:

\begin{equation}
    \mathrm{F}_\beta = (1 + \beta^2) \cdot \frac{\mathrm{recall} \cdot \mathrm{precision}}{\mathrm{recall} + \beta^2 \cdot \mathrm{precision}}.
    \label{eq:f_measure}
\end{equation}

During the experiments, we used $\beta=1$, which puts equal emphasis on both components. This metric is known as F1-score.

\subsection{Pose}
\label{apx:measures:pose}

\subsubsection{PCK (and PCKhn variant)}

The Percentage of Correct Keypoints is defined with some level of tolerance, usually with respect to the bounding box.
In our experiments, we have used a value of \SI{5}{\percent} of the bounding box, denoted as PCK@0.05. We also used a bit more restrictive PCKhn@0.1 variant, which uses \SI{10}{\percent} of hips-neck distance as a tolerance value.

\begin{equation}
    \mathrm{PCK}(\alpha, d) = \frac{1}{N} \sum_{i=1}^{N} 1 \text{~if~} 
        \left\lVert \frac{\mathrm{pred}_{i} - \mathrm{gt}_i}{S} \right\rVert \leq \alpha
    \text{~else~} 0
\label{eq:pck}
\end{equation}

\begin{itemize}
    \item $\alpha$: tolerance level.
    \item $N$: The number of keypoints in the human body pose. Usually, only visible keypoints are included in the calculation, which means that the set can differ from frame to frame.
    \item $i$: Index of a specific keypoint, ranging from 1 to N.
    \item $\mathrm{pred}_{i}$: The predicted coordinates of keypoint $i$ from the pose estimation algorithm. Can be 2D or 3D.
    \item $\mathrm{gt}_{i}$: The ground truth coordinates of keypoint $i$ from the annotated dataset.
    \item $S$: The normalization factor -- usually the bounding box size or a specific body part distance (e.g., hips-neck distance).
\end{itemize}

\subsubsection{MPJPE}

Mean Per-Joint Position Error is used to quantify the error between the predicted joint positions and the ground truth joint positions. It is calculated as the mean Euclidean distance between the corresponding joint positions of the predicted pose and the ground truth pose, usually measured in millimeters.

\begin{equation}
\text{MPJPE} = \frac{1}{N} \sum_{i=1}^{N} \left\lVert \mathrm{pred}_{i} - \mathrm{gt}_i \right\rVert
\label{eq:mpjpe}
\end{equation}

\subsubsection{N-MPJPE}

The Normalized Mean Per-Joint Position Error metric is calculated similarly to MPJPE, but with an additional normalization step to account for variations in scale between different poses. This is especially useful when comparing errors across datasets or subjects with different body sizes.

\begin{equation}
\text{N-MPJPE} = \frac{1}{N} \sum_{i=1}^{N} \left\lVert S \cdot \mathrm{pred}_{i} - \mathrm{gt}_i \right\rVert
\label{eq:n_mpjpe}
\end{equation}

We've used the metric version from \cite{pavllo20193d}, which itself adapts from \cite{rhodin2018unsupervised}.

\subsubsection{PA-MPJPE}
The Procrustes Aligned Mean Per-Joint Position Error metric involves aligning the predicted and ground truth poses using a similarity transformation (translation, rotation, and scaling) before calculating the MPJPE. This alignment removes any global differences between the poses, allowing the evaluation to focus on local joint estimation errors.

We've used the metric version from \cite{pavllo20193d}.

\subsubsection{MPJVE}
MPJVE (Mean Per-Joint Velocity Error) is a metric used to evaluate the performance of human pose estimation algorithms in predicting the velocities of body joints in 3D space. This metric is particularly relevant for applications such as motion capture, where the estimation of joint velocities is essential for understanding human motion.

We've used the metric version from \cite{pavllo20193d}.

\bibliography{bibliography}

\end{document}